\documentclass[11pt]{article}

\usepackage[preprint]{acl}

\usepackage{times}
\usepackage{latexsym}

\usepackage[T1]{fontenc}

\usepackage[utf8]{inputenc}

\usepackage{microtype}

\usepackage{inconsolata}

\usepackage{graphicx}
\usepackage{booktabs}
\usepackage{multirow}
\usepackage{amsmath}
\usepackage{amssymb}

\title{List Counting Failures Are Not One Phenomenon}

\author{
Iyad Ait Hou, Saad Mankarious, Aya Zirikly, and Rebecca Hwa \\
Department of Computer Science \\
The George Washington University \\
Washington, D.C., USA \\
\texttt{\{iyad.aithou, saadm, ayah.zirikly, rebecca.hwa\}@gwu.edu}
}

\begin{document}
\maketitle
\begin{abstract}
Counting the items in a bracketed list looks trivial, yet open-weight chat models often get it wrong.
Prior work usually blames input bottlenecks such as subword fragmentation or attention dilution, which predict that different models should fail in roughly the same way.
Across seven instruct models on identical prompts, however, wrong answers form distinct modes: Qwen and Gemma~27B often flip odd lengths to a nearby even integer, OLMo concentrates errors on a few mid-sized integers, and Llama tends to under-count.
These modes are useful labels rather than a stable family law (Gemma~9B does not reproduce Gemma~27B's odd-to-even drop), and heavier subword fragmentation does not make counting harder on our benchmark.
When the model answers incorrectly, a linear probe can usually still recover the true count from the residual stream.
Matching the same odd-to-even error also does not imply the same late-MLP magnitude fix: scaling a late MLP output helps Qwen modestly but is near null on Gemma~27B under the same protocol, while residual steering can move both only by trading odd gains for even losses.
These results caution against transferring that magnitude fix across models without a transfer check.
\end{abstract}

\section{Introduction}
\label{sec:intro}

Ask seven chat models how many items are in
\texttt{[apple, banana, cherry, \ldots]}, and they often fail even though every item is visible and the task asks only for the list length.
We study \emph{list cardinality}: reporting how many whitespace-separated items appear in a bracketed list (e.g.,
\texttt{[the, quick, brown, fox]} $\to$ $4$).
That is distinct from character counting inside a token (how many \texttt{r}'s in \texttt{strawberry}; \citealp{fu2024struggle,sims2025stochastok}) and from symbolic arithmetic ($37{+}86$; \citealp{nogueira2021investigating,razeghi2022impact,dziri2023faithfate}).
The setting is controlled; a small tool-argument constraint pilot keeps the same odd-to-even drop on Qwen~32B and Gemma~27B (\S\ref{sec:results:protocol}), which motivates studying direct-ask cardinality rather than treating it as a quiz artifact alone.

\begin{figure*}[t]
    \centering
    \includegraphics[width=0.96\textwidth]{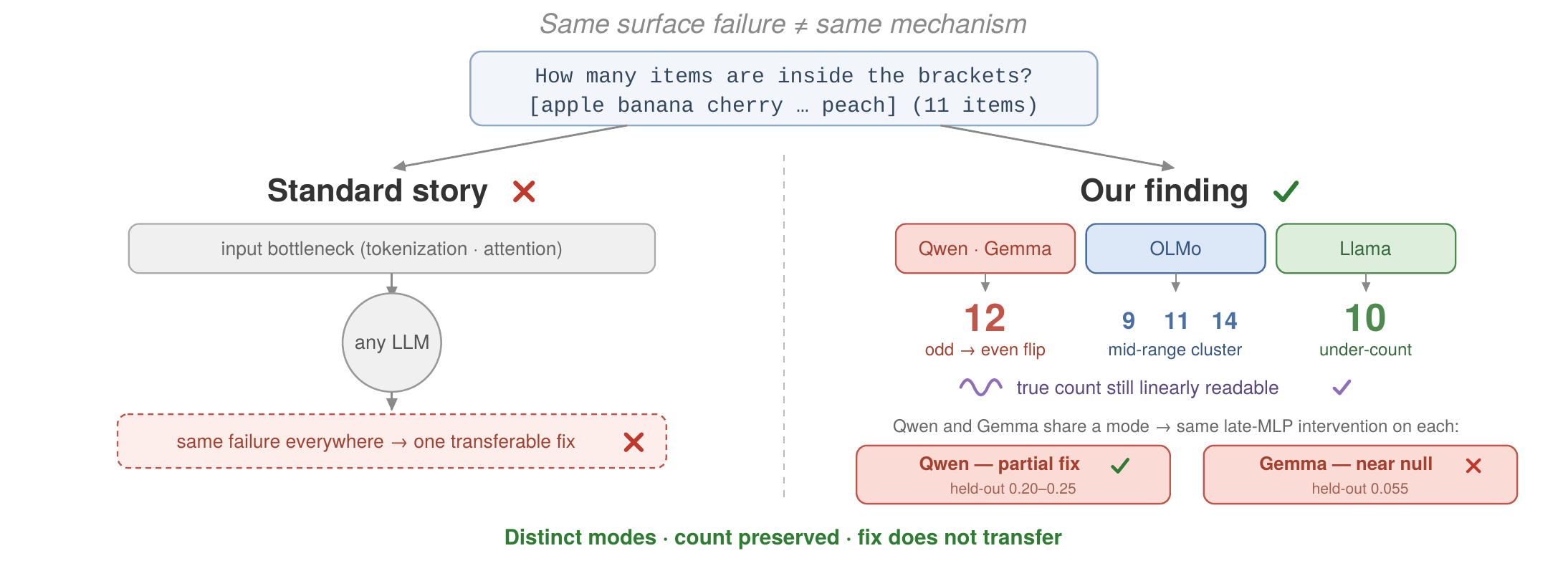}
    \caption{Overview: same surface failure need not mean the same mechanism. The standard story predicts a shared input bottleneck and one transferable late-layer fix. On identical prompts we instead find distinct modes, a usually recoverable residual count when wrong ($0.85$--$0.96$), and non-transfer of a late-MLP magnitude intervention across a shared odd-to-even drop (Qwen $0.20$--$0.25$ vs.\ Gemma~27B $0.055$; Table~\ref{tab:even-odd-gap}; \S\ref{sec:results:probe}--\ref{sec:results:block-decomp}).}
    \label{fig:main-overview}
\end{figure*}

The usual explanations put the problem in the \emph{input}: either tokenization splits items into confusing subwords \citep{singh2024tokenization,zhang2024counting,fu2024struggle,sims2025stochastok}, or attention cannot track every item as the list gets longer \citep{velivckovic2025softmax,nakanishi2025scalable}.
In both stories the model never forms a clean count, so different models should fail in roughly the same way, and heavier tokenization should make things worse.
If that package were right, a late-layer intervention found in one model would be a natural candidate to reuse in another.
Figure~\ref{fig:main-overview} contrasts that shared-bottleneck story with what we find.
We ask four questions on a fixed list-cardinality benchmark:
(i)~Do open-weight chat models fail in one shared way, or in distinct error modes?
(ii)~When the answer is wrong, is the true count missing earlier in the network, or still linearly readable from the residual stream?
(iii)~If two models make the same odd-to-even mistake, do they share a transferable late-MLP \emph{magnitude} fix?
(iv)~Do those modes already appear in base models, or do they strengthen across released post-training stages?

To answer them, we hold the prompts fixed across seven instruct models, compare the shapes of wrong answers, read answer-position residuals with linear probes, and test a late-MLP magnitude intervention on models that share an odd-to-even drop.
Where released stage checkpoints exist, we also compare base and post-training models on the same lattice.
The contribution is that comparison: list-cardinality failure is not one shared phenomenon with one transferable late-layer fix.
We find the following.

\textbf{Error modes differ across models.}
On identical prompts, wrong answers form distinct patterns rather than one shared failure (Fig.~\ref{fig:main-overview}; Table~\ref{tab:even-odd-gap}; \S\ref{sec:results:cross-model}): Qwen and Gemma~27B often answer odd lengths with a nearby even number, OLMo concentrates errors on a few mid-sized integers, and Llama tends to under-count.
These are useful labels, not a stable family law (Gemma~9B does not reproduce Gemma~27B's odd-to-even drop; Appendix~\ref{app:scale-checks}), and heavier BPE fragmentation does not make counting harder (\S\ref{sec:results:cliff}).

\textbf{The true count is usually still readable.}
When the model answers incorrectly, a linear probe recovers the true count at $0.85$--$0.96$ against a permutation baseline near $0.07$ (\S\ref{sec:results:probe}), pointing to late answer failure rather than a missing earlier count.

\textbf{Same surface error need not share a magnitude fix.}
\texttt{reverse\_half} helps Qwen modestly ($0.20$--$0.25$ held-out) but is near null on Gemma~27B ($0.055$) under the same protocol (\S\ref{sec:results:block-decomp}), so this late-MLP magnitude intervention should not be transferred across models without a check.

\textbf{Modes shift across post-training stages.}
Matched Qwen base checkpoints already show an even$-$odd gap that instruction tuning widens (\S\ref{sec:results:protocol}).
On the OLMo~2--32B ladder, base prefers small wrong integers, DPO raises mid-range mass, and instruct makes that concentration clearest.
That is stage sensitivity, not an installing cause: the usable SFT checkpoint refuses with non-integer text, and we still lack a document-level installing statistic.
By ``training family'' we mean pretraining lineage plus tokenizer, chat template, and post-training, not architecture or size alone.

\section{Related work}
\label{sec:related}

\paragraph{Counting in transformers.}
Theory and small-model work characterize when transformers can implement counting and related algorithms \citep{hahn2020theoretical,yehudai2024count,behrens2025counting,golkar2024contextual,chang2025inductive}.
On large LMs, empirical failures are most often attributed to the input encoding: subword tokenization can fragment or obscure the units being counted \citep{singh2024tokenization,zhang2024counting,fu2024struggle,sims2025stochastok}, and softmax attention can lose resolution as the number of items grows \citep{velivckovic2025softmax,nakanishi2025scalable}.
Those accounts predict broadly shared, length- or fragmentation-driven degradation.
They do not predict training-family-specific wrong integers on identical prompts, nor that more fragmented lists can be easier than common-word lists, both of which we observe for list cardinality.
We treat these as constraints on input-side stories for \emph{this} task, not as a blanket refutation of tokenization or attention accounts in other counting settings.

\paragraph{Linear probes and unemitted information.}
A separate literature studies when models encode information that is not reflected in their outputs \citep{burns2022discovering,orgad2024llms,li2023inference}.
Numeric and ordinal structure is often linearly readable from activations \citep{wallace2019nlp,gurnee2024language,heinzerling2024monotonic}, and probing methodology stresses control tasks and selectivity \citep{alain2017understanding,hewitt2019control,belinkov2022probing,giulianelli2018hidden}.
Models also favor high-probability surface forms even when they conflict with the prompt \citep{mccoy2023embers}.
We build on these tools (ridge probes with permutation and transfer controls), but the prior work does not establish that list cardinality remains linearly present under model error, nor that incorrect emissions concentrate on family-dependent integers.
That is an empirical claim of this paper, not a restatement of the probing literature.

\paragraph{Mechanistic analyses of counting.}
Causal and correlational tools have been used to localize numeric computation.
\citet{stolfo2023mechanistic} find late MLP involvement in arithmetic success; logit-lens methods track when a digit becomes readable through the unembedding, with known caveats \citep{nostalgebraist2020logitlens,belrose2023tuned}; \citet{hasani2025counting} study internal counters for repeated-item counting.
Related work on adjacent counting tasks also finds recoverable internal counts with late emission failure, including character counting \citep{earlysuppression2026}, geometric misalignment between count directions and digit unembeddings \citep{wrongdirection2026}, and format-triggered late MLP overwrite on repeated-token lists \citep{repeatedtoken2026}.
Those studies motivate looking past input-side stories, but they do not hold list cardinality fixed across models, compare wrong-integer modes on matched prompts, or test whether a late-layer scaling fix transfers across models that make the same odd-to-even mistake.
Our questions in \S\ref{sec:intro} target that gap.

\section{Experimental setup}
\label{sec:setup}

The experiments follow the four questions in \S\ref{sec:intro}.
We first put many models on the same counting prompts, then ask whether a wrong answer still leaves a readable count inside the network, then test whether a late-MLP magnitude fix transfers, and finally compare released base and post-training stages where those checkpoints exist.
Seeds, pools, and prompt examples are in Appendix~\ref{app:benchmark}.

\subsection{Models}
\label{sec:setup:models}

Table~\ref{tab:models} lists the checkpoints and the role each one plays.
We use open-weight releases from the Qwen~2.5 \citep{yang2024qwen25}, Gemma~2 \citep{gemmateam2024gemma2}, OLMo~2 \citep{olmo2025olmo2}, Llama~3.1 \citep{grattafiori2024llama3}, and DeepSeek-R1 \citep{deepseekai2025r1} model reports, with Mistral Small~3 \citep{mistral2025small} as a held-out family.
The core panel is seven instruct models (Qwen~14B/32B/72B, Gemma~27B, OLMo~32B, Llama~70B, R1-Distill): each gets the main behavioral eval and an answer-position probe.
Gemma~9B and Llama~8B are smaller within-family checks on the same prompt lattice, used only behaviorally (Appendix~\ref{app:scale-checks}).
Mistral-Small-24B is held out for one predictivity test of the late-MLP magnitude intervention (\S\ref{sec:results:protocol}).

For question~(iv) we also evaluate released stage checkpoints on the same \textsc{common\_words} lattice (\S\ref{sec:results:protocol}).
For Qwen~14B/32B that means matched base vs.\ instruct.
For OLMo~2--32B we use the released post-training ladder (base, DPO, instruct); the released SFT checkpoint is excluded because it almost always refuses with non-integer text.
We additionally report two negative installing checks for OLMo (a $50$k-document mix integer-frequency proxy and chosen-vs-rejected integer mining in the preference mix), without claiming an installing cause.
The main $1{,}300$-prompt cross-model comparison remains instruct-only.

\begin{table}[t]
\centering
\footnotesize
\setlength{\tabcolsep}{3pt}
\resizebox{\columnwidth}{!}{%
\begin{tabular}{@{}llc@{}}
\toprule
Model & Role & Prec. \\
\midrule
Qwen~2.5--14B & within-family scale & bf16 \\
Qwen~2.5--32B & reference & bf16 \\
Qwen~2.5--72B & within-family scale & nf4 \\
Gemma~2--9B & within-family scale & bf16 \\
Gemma~2--27B & cross-family & bf16 \\
OLMo~2--32B & cross-family (+ stages) & bf16 \\
Llama~3.1--8B & within-family scale & bf16 \\
Llama~3.1--70B & cross-family & nf4 \\
R1-Distill-Qwen-32B & reasoning distill & bf16 \\
Mistral-Small-24B & OOD lever check & bf16 \\
\bottomrule
\end{tabular}%
}
\caption{Models. Core seven: main eval + probes; Gemma~9B / Llama~8B: scale checks (Appendix~\ref{app:scale-checks}); Mistral-Small-24B: held-out \texttt{reverse\_half} test. Stages: Qwen base vs.\ instruct; OLMo base$\to$DPO$\to$instruct (\S\ref{sec:results:protocol}).}
\label{tab:models}
\end{table}

\subsection{Task and data}
\label{sec:setup:data}

To compare error modes fairly, every model must see the same prompts.
The task is list cardinality: given a whitespace-separated list in brackets, return the number of items.
The main eval uses one English instruction, wrapped in each model's chat template:
\texttt{How many items are inside the brackets? Return only one integer. Items: [\{list\}]}.
We keep the ask fixed and vary only list surface form, so cross-model differences are not confounded by different questions.
Alternate English paraphrases are a later robustness check (\S\ref{sec:results:protocol}). Multilingual lists are surface-form controls under that fixed English ask, not a test of multilingual counting competence.

For each length $L\in\{3,\ldots,15\}$ and each of five surface conditions (Table~\ref{tab:conditions}), we sample $20$ lists from a closed pool with a fixed seed.
That yields $1{,}300$ prompts; every model sees the identical strings.
The conditions pressure input-side stories: \textsc{common\_words} is the familiar baseline, \textsc{rare\_words} lowers frequency, the multilingual rows change script/surface, and \textsc{random\_strings} multiplies tokens per item by about $4\times$ as a tokenization-load control (items are also highly distinctive, so this does not rule out every input-side account; \S\ref{sec:results:cliff}).
We decode greedily, take the first integer in the assistant turn (after \texttt{</think>} for R1-Distill), and report accuracy, the even$-$odd gap on lengths $\{10,12,14\}$ vs.\ $\{11,13,15\}$ with bootstrap $95\%$ CIs ($300$ prompts per side), and the distribution of wrong integers.

\begin{table}[t]
\centering
\footnotesize
\setlength{\tabcolsep}{3pt}
\resizebox{\columnwidth}{!}{%
\begin{tabular}{@{}lll@{}}
\toprule
Condition & Varies & Purpose \\
\midrule
\textsc{common\_words} & high-freq.\ English & Familiar baseline \\
\textsc{rare\_words} & rare English & Frequency control \\
\textsc{ml\_single} & one non-English lang. & Script/surface OOD \\
\textsc{ml\_mixed} & mixed scripts & Script-mixing stress \\
\textsc{random\_strings} & alphanumeric pseudowords & Tokenization load \\
\bottomrule
\end{tabular}%
}
\caption{Surface conditions under a fixed English ask. \textsc{ml\_*} $=$ multilingual; multilingual rows stress list surface form, not multilingual counting competence. Examples in Appendix~\ref{app:benchmark}.}
\label{tab:conditions}
\end{table}

\subsection{Probing}
\label{sec:setup:probe}

Behavioral modes alone do not tell us whether the count is missing inside the network.
For every core model we therefore cache the answer-position residual on an expanded \textsc{common\_words} set ($40$ lists per length; $520$ prompts) and fit a ridge regressor ($\alpha{=}1$, $5$-fold CV) from residual to integer count at each layer.
We report round-accuracy with permutation and leave-one-count-out controls, a matched probe trained on the emitted integer, and (on Qwen~32B) zero-shot transfer to other surface residuals (\S\ref{sec:results:probe}).

\subsection{Controls and interventions}
\label{sec:setup:interventions}

Before intervening, we check two simple behavioral alternatives to a shared input bottleneck: whether \textsc{random\_strings} is harder than \textsc{common\_words} on all seven models, and whether Qwen~32B answers the item count or the whitespace-chunk count when those differ ($n{=}40$ per $N$).
A filler-padding check on Qwen~32B is appendix-only and is not used as a dilution refutation (Appendix~\ref{app:interventions}).

The causal analysis then asks whether models that make the same odd-to-even mistake share a late-MLP magnitude handle.
We follow a fixed discovery chain rather than an open search: logit lens points to a late flip, block decomposition attributes most of the wrong-direction update to the MLP, and a small per-failure sweep nominates scaling that MLP output by $-0.5$ (\texttt{reverse\_half}).
We report selection-corrected held-out fix rates over all probe-cell odd-length failures for Qwen~32B, Qwen~14B, and Gemma~27B ($20$ split-half rounds).
Qwen~32B also includes class-mean steering and a global scale sweep that exposes the even/odd trade-off.
Details are in Appendix~\ref{app:interventions}.

\section{Results}
\label{sec:results}

We start with behavior on the shared prompt lattice, then ask what remains readable when the answer is wrong, then whether a late-MLP magnitude fix transfers, and finally how modes change across released post-training stages.
Unless noted, main-eval numbers use identical prompts across models, greedy decoding, and the even$-$odd gap on lengths $\{10,12,14\}$ vs.\ $\{11,13,15\}$ ($300$ prompts/side; bootstrap $95\%$ CIs). Full grids are in Appendices~\ref{app:full-grid}--\ref{app:per-model-grid}.

\subsection{Error modes differ across models}
\label{sec:results:cross-model}

The first question is whether failures look alike.
On identical prompts they do not: wrong answers form distinct modes rather than one shared length-driven pattern.
Table~\ref{tab:even-odd-gap} summarizes the panel: Qwen~14B/32B, R1-Distill, and Gemma~27B show a large even$-$odd gap (parity drop; $+0.24$ to $+0.40$); Qwen~72B weakens it ($+0.10$) and shifts top wrongs off the even neighbors; OLMo has no reliable gap (CI crosses zero) and concentrates wrongs on a few mid-range integers rather than flipping parity; Llama~70B inverts the pooled gap and under-counts.
We treat ``mode class'' as an operational label from three observables (even$-$odd gap (sign/CI), $P(\hat{y}{<}y\mid$ wrong$)$, and top-wrong mass), not as a claim that any particular wrong integer is lineage-stable: OLMo/Llama both under-count ($0.95$--$0.99$) while Qwen~32B's parity wrongs are mixed ($0.46$), which separates under-count from parity-flip even when their top-wrong lists overlap.
Pooled gaps ($n{=}300$/side) carry these claims; Qwen~32B adjacent cells likewise separate (e.g.\ $L{=}10$: $1.00$ vs.\ $L{=}11$: $0.05$; Appendix~\ref{app:full-grid}), with directional signed errors (Table~\ref{tab:full-signed}).
Per-condition length curves are in Appendix~\ref{app:visual-diagnostics}.
Architecture and scale alone do not determine the shape: same-scale OLMo lacks Gemma~27B's drop, and R1 distillation does not remove Qwen's.
Within-family scale checks (Appendix~\ref{app:scale-checks}) support a Llama under-count class at 8B (bf16) and 70B (nf4; descriptive).
Gemma~9B does \emph{not} reproduce Gemma~27B's odd-to-even drop: it collapses onto a fixed mid-range integer for longer lists, which makes some odd lengths accidentally correct and drives the even$-$odd gap negative ($-0.30$ vs.\ $+0.25$).
Lineage often predicts mode class better than matched scale (OLMo vs.\ Qwen at 32B), but checkpoints within a lineage can shift class.

\begin{table*}[t]
\centering
\small
\setlength{\tabcolsep}{3.2pt}
\begin{tabular}{@{}lccccccc@{}}
\toprule
Model & Gap $[95\%$ CI$]$ & Shape & Acc$_{\mathrm{c}}$ & Acc$_{\mathrm{r}}$ & $\Delta$ & $P($u$\mid$w$)$ & Top wrong \\
\midrule
Qwen-32B & $+0.40$ $[0.33,0.47]$ & parity & 0.70 & 0.92 & $+0.22$ & 0.46 & 12, 14, 8 \\
R1-Distill & $+0.39$ $[0.32,0.46]$ & parity & 0.45 & 0.70 & $+0.24$ & 0.92 & 12, 6, 8 \\
Qwen-14B & $+0.36$ $[0.29,0.44]$ & parity & 0.58 & 0.77 & $+0.19$ & 0.75 & 12, 13, 8 \\
Gemma-27B & $+0.24$ $[0.18,0.30]$ & parity & 0.38 & 0.46 & $+0.08$ & 0.94 & 10, 11, 12 \\
Qwen-72B & $+0.10$ $[0.02,0.18]$ & weak parity & 0.86 & 0.93 & $+0.07$ & 0.11 & 13, 14, 15 \\
OLMo-32B & $+0.05$ $[-0.01,0.12]$ & mid-range & 0.30 & 0.45 & $+0.16$ & 0.95 & 11, 8, 9 \\
Llama-70B & $-0.24$ $[-0.31,-0.17]$ & under-count & 0.43 & 0.54 & $+0.11$ & 0.99 & 11, 10, 12 \\
\bottomrule
\end{tabular}
\caption{Error modes and tokenization control on identical prompts.
Gap $=$ even$-$odd accuracy ($\{10,12,14\}$ vs.\ $\{11,13,15\}$; $300$/side; bootstrap $95\%$ CIs).
Acc$_{\mathrm{c}}$ / Acc$_{\mathrm{r}}$ $=$ mean accuracy on \textsc{common\_words} / \textsc{random\_strings}; $\Delta{=}$Acc$_{\mathrm{r}}{-}$Acc$_{\mathrm{c}}$ (random uses ${\sim}4\times$ more tokens at $L{=}15$).
$P($u$\mid$w$)$ $=$ $P(\hat{y}{<}y\mid$ wrong$)$ on \textsc{common\_words} (separates mid-range concentration from pure under-count).
Top wrong $=$ leading wrong emissions on \textsc{common\_words} (lattice-specific; not a length-invariant attractor).
Full grids in Appendices~\ref{app:full-grid}--\ref{app:per-model-grid}.}
\label{tab:even-odd-gap}
\label{tab:tokenization-control}
\end{table*}

\subsection{Heavier tokenization is not harder on this benchmark}
\label{sec:results:cliff}

A natural follow-up is whether those failures are just tokenization load.
If subword fragmentation caused them, \textsc{random\_strings} should be hardest.
Table~\ref{tab:tokenization-control} shows the opposite on our lattice: Acc$_{\mathrm{r}}>$Acc$_{\mathrm{c}}$ for every model ($\Delta$ from $+0.07$ to $+0.24$) despite ${\sim}4\times$ more tokens per item at $L{=}15$.
Items are also more distinctive, so this rules out ``more tokens $\Rightarrow$ harder'' without closing every input-side story (Limitations).
A multi-word-item control on Qwen~32B likewise fails a chunk-counting prediction: at even $N$, answers favor the item count ($\sim$65--78\%) over the whitespace-chunk count ($\le$12.5\%), and at odd $N$ they fall onto even integer modes (e.g.\ $85\%$ say ``12'' at $N{=}11$) rather than either count.
Parity oscillation is also hard to reconcile with monotone attention-dilution stories; we do not treat filler-padding as a causal dilution refutation (Appendix~\ref{app:interventions}).
The breaks are therefore better read as model-conditioned answer structure than as a generic input bottleneck.

\subsection{The true count usually survives when the answer is wrong}
\label{sec:results:probe}
\label{sec:results:probe-cross-model}
\label{sec:results:logit-lens}

Modes and tokenization controls still leave open whether the count is missing inside the network.
When greedy decoding is wrong, a linearly recoverable true-count signal is usually still present at the answer position.
Table~\ref{tab:cross-model-probe} shows last-layer probe round-accuracy near $0.92$--$0.98$ across all seven models, while model accuracy ranges from roughly $0.30$ to $0.89$ on the same \textsc{common\_words} prompts.
The raw ``probe right, model wrong'' counts track the independence product $\text{probe acc.}\times(1-\text{model acc.})$ within $0.02$ for every model, so they are not themselves evidence of a special residual; the conditional controls in Table~\ref{tab:probe-controls} carry the claim.
On model-wrong subsets, true-count probe accuracy remains $0.85$--$0.96$ against a permuted-label baseline near $0.07$ and a majority-class baseline of $0.11$--$0.26$ (always predict the modal true count among wrongs); a matched probe trained on the emitted integer reaches only $0.40$--$0.76$ on the same residuals.
The contrast is cleanest for Qwen and weaker for OLMo/Llama, where wrong emissions are more concentrated (Limitations).
Read-only diagnostics agree on timing: at $L{=}11$, after forcing the leading ``1'', Qwen~32B assigns $P(\text{``2''})=0.885$ vs.\ $P(\text{``1''})=0.111$, and logit-lens traces show mid-network correct-digit mass overwritten late (Appendix~\ref{app:visual-diagnostics}).
We read this as late emission failure with a preserved residual count, not as proof that the frozen unembedding uses the probe's coordinates; the probe interpolates within the trained range but does not extrapolate beyond it (Appendix~\ref{app:probe-extrap}), consistent with a bounded-resolution count axis rather than an unbounded cardinality direction.

\begin{table*}[t]
\centering\small
\setlength{\tabcolsep}{4pt}
\begin{tabular}{lrrrrr}
\toprule
model & $n$ & model acc.\ & probe acc.\ & probe right, model wrong & \% at $\{11,13,15\}$ \\
\midrule
OLMo 2--32B & 520 & 0.302 & 0.962 & 346 (66.5\%) & 32\% \\
R1-Distill Qwen 32B & 520 & 0.402 & 0.935 & 290 (55.8\%) & 33\% \\
Gemma 2--27B & 520 & 0.392 & 0.919 & 285 (54.8\%) & 35\% \\
Llama 3.1--70B (4-bit) & 520 & 0.465 & 0.979 & 268 (51.5\%) & 43\% \\
Qwen 2.5--14B & 520 & 0.575 & 0.927 & 197 (37.9\%) & 47\% \\
Qwen 2.5--32B & 520 & 0.704 & 0.923 & 131 (25.2\%) & 69\% \\
Qwen 2.5--72B (4-bit) & 520 & 0.887 & 0.958 & 51 (9.8\%) & 27\% \\
\bottomrule
\end{tabular}
\caption{Cross-model probe-vs-model accuracy on the same $520$-sequence \textsc{common\_words} cell ($L\!\in\!\{3,\ldots,15\}$, $40/L$). ``Probe right, model wrong'' counts sequences where the linear probe at $\ell_{\text{last}}$ predicts the correct integer while the model emits the wrong digit. The last column reports the share of those cases at odd lengths $\{11,13,15\}$.}
\label{tab:cross-model-probe}
\end{table*}

\begin{table*}[t]
\centering
\footnotesize
\setlength{\tabcolsep}{3.5pt}
\resizebox{\textwidth}{!}{%
\begin{tabular}{lrrrrrr}
\toprule
 & & & \multicolumn{2}{c}{true-count probe} & perm. & emit-probe \\
\cmidrule(lr){4-5} \cmidrule(lr){7-7}
model & model acc. & probe acc. & $P(\checkmark\mid\times)$ & $P(\checkmark\mid\checkmark)$ & round-acc & on errors \\
\midrule
Qwen 2.5--32B          & 0.704 & 0.923 & 0.851 {\scriptsize[0.79, 0.90]} & 0.954 & 0.071 & 0.448 \\
Qwen 2.5--14B          & 0.575 & 0.927 & 0.891 {\scriptsize[0.84, 0.93]} & 0.953 & 0.066 & 0.552 \\
Qwen 2.5--72B (4-bit)  & 0.887 & 0.958 & 0.864 {\scriptsize[0.76, 0.93]} & 0.970 & 0.073 & 0.441 \\
Gemma 2--27B           & 0.392 & 0.919 & 0.902 {\scriptsize[0.86, 0.93]} & 0.946 & 0.066 & 0.665 \\
OLMo 2--32B            & 0.302 & 0.962 & 0.953 {\scriptsize[0.93, 0.97]} & 0.981 & 0.076 & 0.730 \\
R1-Distill Qwen 32B    & 0.402 & 0.935 & 0.933 {\scriptsize[0.90, 0.96]} & 0.938 & 0.070 & 0.399 \\
Llama 3.1--70B (4-bit) & 0.465 & 0.979 & 0.964 {\scriptsize[0.94, 0.98]} & 0.996 & 0.070 & 0.763 \\
\bottomrule
\end{tabular}%
}
\caption{Probe controls on the $520$-sequence \textsc{common\_words} cell, last-layer answer-position residual, all seven models. $P(\checkmark\mid\times)$ / $P(\checkmark\mid\checkmark)$: held-out true-count probe round-accuracy on model-wrong / model-correct subsets (Wilson $95\%$ CIs). \emph{perm.}: mean round-accuracy under five label permutations (chance ${\approx}0.077$). \emph{emit-probe on errors}: probe trained on the model's emitted integer, evaluated on model-wrong residuals. On the same residuals where the true count is decodable at $0.85$--$0.96$, the wrong emission is decodable at only $0.40$--$0.76$.}
\label{tab:probe-controls}
\end{table*}

\subsection{A late-MLP scale lever helps Qwen, not Gemma}
\label{sec:results:block-decomp}
\label{sec:results:global-l52}
\label{sec:results:feature}

\begin{figure*}[t]
    \centering
    \includegraphics[width=0.96\textwidth]{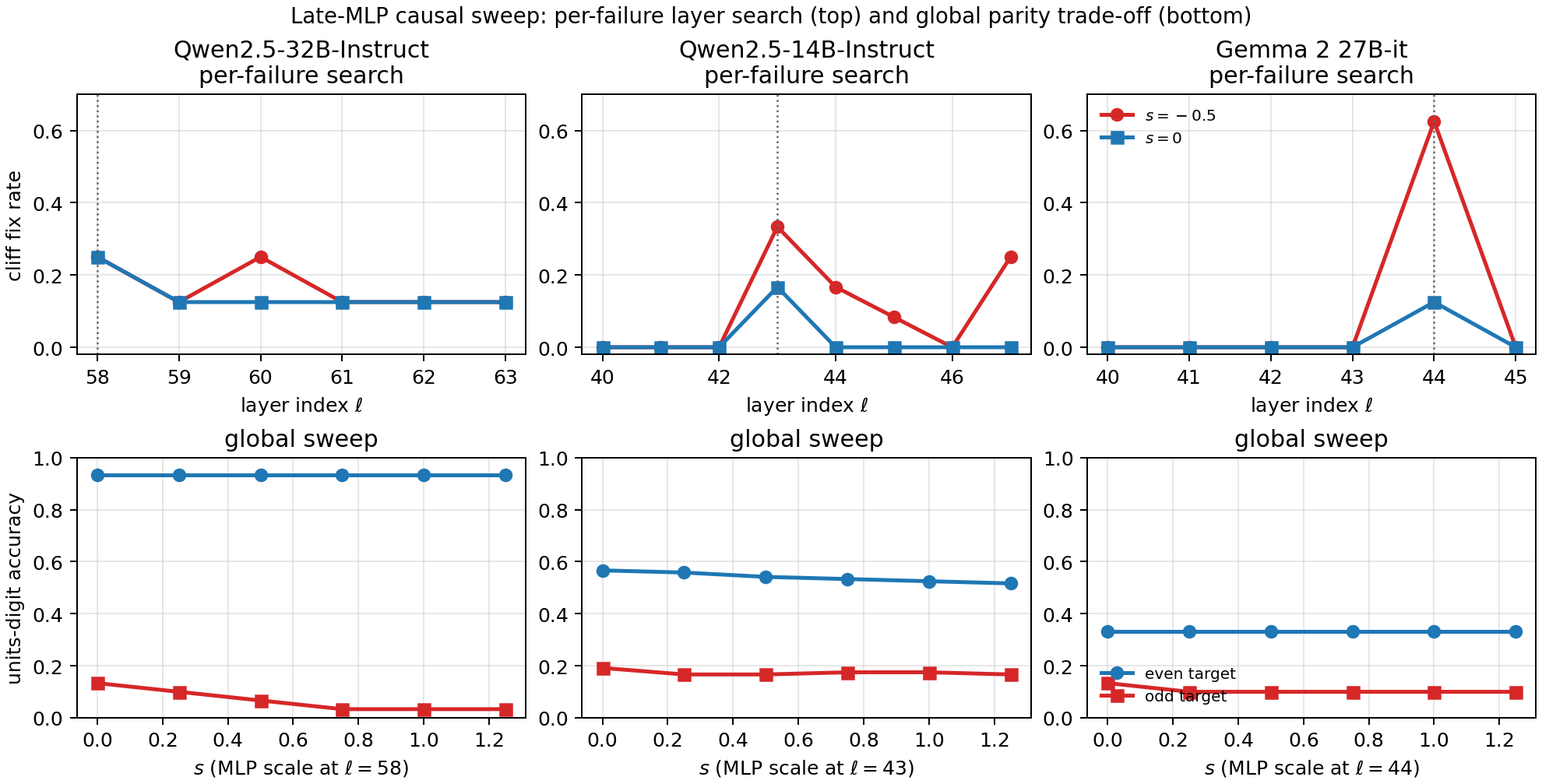}
    \caption{Causal MLP scale interventions. \textbf{Top:} per-layer fraction of odd-length failures fixed by scaling that layer's MLP output by $s\in\{-0.5,0\}$ over the late-layer candidate set (Appendix~\ref{app:interventions}; small-sample panels are upper bounds). Selection-corrected full-failure held-out rates: $0.20\pm0.06$ (Qwen~32B), $0.25\pm0.05$ (Qwen~14B), $0.055\pm0.022$ (Gemma~27B). \textbf{Bottom:} even- vs.\ odd-target accuracy as the MLP output scale $s$ sweeps on Qwen~32B ($s{=}1$ is baseline). Layer indices differ by analysis (discovery vs.\ split-half vs.\ continuous sweep); see Appendix~\ref{app:interventions}.}
    \label{fig:cross-model-causal}
\end{figure*}

If two models make the same odd-to-even mistake, it is natural to reuse the same late-layer fix.
We test that transfer for one concrete intervention: scaling a late MLP output (\texttt{reverse\_half}).
The informative result is not the modest fix rate but its zero-sum structure: the same surface drop does not imply a shared magnitude handle.
Figure~\ref{fig:cross-model-causal} summarizes the full-failure protocol: selection-corrected held-out recovery is $0.20\pm0.06$ on Qwen~32B and $0.25\pm0.05$ on Qwen~14B, but only $0.055\pm0.022$ on Gemma~27B, despite Gemma's matching odd-to-even drop and high last-layer probe (${\sim}0.92$).
Within Qwen, a continuous late-MLP scale sweep (module output ${\times}s$; $s{=}1$ is the unsteered baseline) raises even-target accuracy from $0.755$ to $0.865$ while odd-target accuracy falls from $0.653$ to $0.581$, and \texttt{reverse\_half} fixes $84$ cases while breaking $182$.
Class-mean residual steering can move all three models with that drop (best cliff-fix ${\sim}0.33$ / $0.43$ / $0.28$ on Qwen~32B / Qwen~14B / Gemma~27B), but only as a zero-sum trade-off (Gemma correct-control accuracy $1.0\to 0.35$).
What fails to transfer is therefore the MLP-\emph{magnitude} lever, not every residual intervention; small-sample vignettes overstated both families (Qwen $6/8$; Gemma ${\approx}0.69$ on $n{=}8$), and the full-failure protocol is the claim.
This is consistent with a small late write being geometrically cheap: adjacent digit unembeddings are highly aligned (most aligned pair $(1,2)$ at $\cos{=}0.865$; Appendix~\ref{app:visual-diagnostics}).
The reading is a miscalibrated prior write in Qwen that magnitude scaling can partially reverse, and a different usable direction in Gemma: matching surface mode, non-matching magnitude lever.

\subsection{Developmental and protocol analyses}
\label{sec:results:protocol}

The last question is whether modes are already present before instruction tuning, or strengthen across released stages; we also check how fragile they are to how the count is asked.
On Qwen~14B/32B, matched base checkpoints already show a \textsc{common\_words} even$-$odd gap (bootstrap $95\%$ CIs exclude zero: $0.47$ $[0.33,0.60]$ at 32B; $0.22$ $[0.05,0.38]$ at 14B), and instruction tuning widens those gaps to $0.77$ and $0.53$.
Wrong-emission mass on the Qwen even modes $\{8,12,14\}$ likewise rises under instruct ($0.45{\to}0.72$ at 32B; $0.45{\to}0.65$ at 14B; endpoint CIs exclude each other).
For OLMo~2--32B we evaluate the released post-training ladder on the same \textsc{common\_words} lattice: base lacks an odd-to-even drop and, when wrong, prefers small integers ($5$/$6$/$4$; modal-wrong mass on a mid-range integer $0.07$); the intermediate DPO stage raises that mass to $0.17$ (top wrongs still mixed); the final instruct checkpoint reaches $0.28$ with that mid-range peak clearest on this lattice.
The released SFT checkpoint is not usable here: almost all generations are non-integer ``write a Python solution\ldots'' refusals, so we exclude it from the mass series.
A $50$k-document OLMo-mix integer-frequency proxy correlates with accuracy overall ($r{=}0.66$) but not for $N\ge8$ ($r{=}{-}0.12$); mining chosen-vs-rejected integers in \texttt{allenai/olmo-2-0325-32b-preference-mix} likewise yields no overweight of that mid-range integer.
A wrong-neighbor emission-mass proxy is stronger for Qwen~32B ($r{=}{-}0.86$) than for OLMo ($r{=}{-}0.59$).
We therefore treat the OLMo concentration as late-post-training sensitive rather than a pretrain parity prior, without claiming an installing document statistic or a single causal stage.

Modes are also protocol-conditioned.
Ask paraphrases (\texttt{count\_entries}, \texttt{how\_many}, \texttt{cardinality}; $20$ lists $\times$ lengths $\{10,\ldots,15\}$) leave the odd-to-even drop intact: Gemma~27B gaps stay in $[0.27,0.33]$, Qwen~14B in $[0.38,0.52]$, Qwen~32B in $[0.80,0.92]$.
Forced enumeration before the integer (\texttt{enumerate\_then\_answer}) abolishes direct-ask failures where we tested it: Qwen~32B and Gemma~27B both reach ${\approx}0.98$--$1.00$ even and odd accuracy (gap $0$; $n{=}120$); OLMo~32B likewise jumps from floor under ask paraphrases to even $1.00$ / odd $0.97$ (gap ${\approx}0.03$; $n{=}120$).
We did not run the enumeration template on R1-Distill; R1's retained drop under the default ask is already evidence that reasoning distillation alone does not remove the mode.
As a minimal ecological check, the same item lists under a tool-argument constraint (``output the integer number of arguments you will pass'') preserve the drop on both Qwen~32B (gap $0.90$ vs.\ $0.88$ on the bracket control; $n{=}120$) and Gemma~27B (gap $0.20$ vs.\ $0.27$); odd-length wrongs remain nearby even integers.
A warehouse-inventory paragraph frame is mixed: Gemma keeps a gap ($0.23$), while Qwen collapses on both even and odd lengths (gap $0.13$ with even accuracy only $0.13$). Prose embedding can change absolute difficulty without removing Gemma's drop or Qwen's tool-constraint drop.
As a held-out concentration$\to$\texttt{reverse\_half} check, Mistral-Small-24B-Instruct-2501 shows a parity cliff (gap ${\approx}0.28$) with diffuse neighbor mass (${\approx}0.37$), so the pre-registered rule predicts near-null recovery, yet selection-corrected held-out fix rate is $0.28\pm0.06$ (a miss).
Concentration alone does not forecast the lever; boundaries for interpretation are collected in Limitations.

\section{Discussion}
\label{sec:discussion}

The results relocate list-cardinality failure from a shared input bottleneck to family-conditioned late emission.
Tokenization-load and monotone dilution accounts predict shared, input-driven degradation \citep{singh2024tokenization,zhang2024counting,velivckovic2025softmax,nakanishi2025scalable}; family-conditioned modes, an inverted surface-form ordering, and a surviving residual count jointly push against that package on this task.
None of this proves the frozen unembedding reads the probe's coordinates.

We attribute the split to family-level factors---pretraining lineage, tokenizer, template, and post-training---not architecture or scale alone.
Modes strengthen across released stages, survive a tool-argument framing, and vanish under forced enumeration: they behave like emission biases \citep{mccoy2023embers}, not a missing counting algorithm.
We have not identified the installing training signal.

Mechanistically, what is ``wrong in the model'' is not one shared late-MLP overwrite.
Qwen has a partially reversible late MLP prior write (zero-sum under scale); Gemma shares the surface cliff and a residual steering handle, but not that MLP magnitude lever (\S\ref{sec:results:block-decomp}).
Read-only diagnostics can show mid-network correct-digit mass in both families without implying that the bad update is an MLP output you can reverse-scale.
Concentration of wrong mass is at best an in-sample correlate of \texttt{reverse\_half} usefulness, not a law (\S\ref{sec:results:protocol}).
Together with the aligned digit geometry and bounded residual count axis reported in \S\ref{sec:results:probe}--\ref{sec:results:block-decomp}, the Qwen odd-length collapse reads as an ordinally brittle output head plus a late family-local write.

\section{Conclusion}
\label{sec:conclusion}

On list cardinality, counting failure is not one phenomenon.
Lineage-conditioned integer modes appear on identical prompts, survive ask paraphrases, and (for Qwen/Gemma) survive a tool-argument constraint framing; stated tokenization-load and chunk-counting accounts fail their tests; a true-count signal is usually still linearly recoverable when emission is wrong; and models that share a parity cliff need not share a late-MLP \emph{magnitude} lever (Qwen recovers $0.20$--$0.25$ under \texttt{reverse\_half} while Gemma is near null; residual steering can still move both, only zero-sum).
Forced enumeration abolishes the Qwen~32B, Gemma~27B, and OLMo~32B direct-ask gaps; R1-Distill retains a parity cliff under the default ask, evidence that the mode is emission/protocol-conditioned on this benchmark.
Error shape remains a first-class object for transfer checks: a simple concentration$\to$lever predictor misses on size-matched Mistral-Small-24B.

\section*{Limitations}
\label{sec:limitations}

The story this paper tells is about a specific ask: an English, no-CoT request for the length of a bracketed list.
Ask paraphrases leave the Qwen and Gemma cliffs intact, and a tool-argument constraint preserves them on Qwen~32B and Gemma~27B, so the failure is not only a bracket-quiz artifact; forced enumeration abolishes the same gaps (\S\ref{sec:results:protocol}).
Mid-generation and multi-turn settings remain out of scope.

On the representation side, a high probe score means the count is linearly present in the residual, not that the frozen unembedding reads that direction.
The true-count probe still beats permutation and majority-class baselines on model-wrong subsets, but the emit-versus-true contrast is weaker for OLMo and Llama than for Qwen and Gemma, so the cleanest ``knows but won't say'' reading is family-conditioned rather than universal.

The late-MLP magnitude lever helps Qwen and is near-null on Gemma; residual steering can still move both, but only as a zero-sum trade.
Wrong-mass concentration looked like a predictor of that lever in-sample and then missed on size-matched Mistral-Small-24B, so we treat concentration as a correlate, not a transfer rule.

Finally, what we call lineage mixes tokenizer, chat template, and post-training, so we cannot say which of those sets the mode class---and that class can even change within one line (Gemma~9B; Appendix~\ref{app:scale-checks}).
We also never found a training-document count that explains why those attractor integers win.
Llama~70B and Qwen~72B results are nf4 only (descriptive); multi-word and filler checks are Qwen-32B-only; and with $n{=}20$ per cell, the main claims use pooled contrasts.


\bibliography{custom}

\section*{Acknowledgments}
Code and figure drafting used AI coding assistants; all reported numbers come from the released scripts and CSVs. The authors reviewed and take responsibility for the final text and results.

\appendix

\section*{Contents of the appendix}
\noindent
\textit{A}~\S\ref{app:benchmark}: benchmark construction and examples.\\
\textit{B}~\S\ref{app:scale-checks}: within-family scale checks.\\
\textit{C}~\S\ref{app:full-grid}: Qwen~32B accuracy and signed-error grids.\\
\textit{D}~\S\ref{app:per-model-grid}: remaining models' accuracy grids.\\
\textit{E}--\textit{F}~\S\ref{app:probe-extrap}--\ref{app:probe-per-length}: probe extrapolation and per-length detail.\\
\textit{G}~\S\ref{app:interventions}: causal intervention protocols.\\
\textit{H}~\S\ref{app:tokenization-control}: tokenization-control note.\\
\textit{I}~\S\ref{app:error-fingerprint}: error-mode separability.\\
\textit{J}~\S\ref{app:visual-diagnostics}: additional figures.\\
\textit{K}~\S\ref{app:reproducibility}: seeds, compute, and release.

\section{Benchmark construction and examples}
\label{app:benchmark}

This is a controlled synthetic benchmark, not a scraped corpus.
Every list is generated by sampling items from a fixed pool with
\texttt{numpy.random.default\_rng(seed)} (Table~\ref{tab:repro-seeds}), then joining them with spaces inside the shared instruction template of \S\ref{sec:setup:data}.
Sampling is without replacement when $L$ is at most the pool size, and with replacement otherwise.
The same seeds (and therefore the same $1{,}300$ strings) are used for every model.

\paragraph{Pools.}
\textsc{common\_words}: $48$ frequent English nouns (e.g.\ \texttt{apple}, \texttt{table}, \texttt{window}).
\textsc{rare\_words}: $40$ low-frequency English words (e.g.\ \texttt{perspicacious}, \texttt{sesquipedalian}).
\textsc{multilingual\_single}: $20$ everyday nouns each in Spanish, Russian, Chinese, and Arabic; each list is drawn from one language.
\textsc{multilingual\_mixed}: the union of those four lexicons in one pool.
\textsc{random\_strings}: $200$ pseudowords of length $4$--$6$ over consonants and digits (seed $303$), built to fragment under BPE.

\paragraph{Design scope.}
The benchmark is a controlled stress test with fixed seeds, closed pools, identical prompts across models, and a tokenization condition that multiplies tokens per item by ${\sim}4\times$. It is not a naturally occurring corpus with human annotation or ecological coverage of how people ask counting questions.
$n{=}20$ per cell supports the even--odd aggregate ($n{=}300$/side) but is thin for single-cell point estimates; we therefore lean on gaps and CIs in the main text.

\begin{table*}[t]
\centering
\small
\setlength{\tabcolsep}{4pt}
\begin{tabular}{@{}llp{0.72\textwidth}r@{}}
\toprule
Condition & $L$ & Example list (items only) & Tok \\
\midrule
\textsc{common\_words} & 5 & beach door window letter bird & 5 \\
\textsc{common\_words} & 11 & water office car fish friend story rock fire road forest phone & 11 \\
\textsc{rare\_words} & 11 & truculent ineffable quotidian \ldots sesquipedalian & 31 \\
\textsc{ml\_single} & 11 & Arabic nouns (e.g.\ school, garden, \ldots market) & 23 \\
\textsc{ml\_mixed} & 11 & mixed Chinese/Spanish/Arabic/Russian nouns & 21 \\
\textsc{random\_strings} & 11 & 7nyh5 wtcp 88h8jc 3bwrz \ldots pvkp5 & 49 \\
\bottomrule
\end{tabular}
\caption{Example lists from the released eval (Qwen tokenizer token counts at right). Same strings are shown to every model; multilingual rows are abbreviated here for layout (full Unicode lists ship in the CSV release).}
\label{tab:benchmark-examples}
\end{table*}

\section{Within-family scale checks}
\label{app:scale-checks}

Gemma~9B and Llama~8B are evaluated on the same $1{,}300$-prompt lattice as the core panel (main-eval only; no second probe/causal stack).
Table~\ref{tab:scale-checks} reports \textsc{common\_words} even$-$odd gaps.
Llama~8B matches the Llama~70B under-count class.
Gemma~9B does not reproduce Gemma~27B's odd-to-even drop: its negative gap comes from collapsing onto a fixed mid-range integer on longer lists, so some odd true lengths are accidentally correct.

\begin{table*}[t]
\centering
\footnotesize
\setlength{\tabcolsep}{4pt}
\resizebox{\textwidth}{!}{%
\begin{tabular}{@{}lcccc@{}}
\toprule
Model & Gap$_{cw}$ $[95\%$ CI$]$ & Shape & $P(\mathrm{u}\mid\mathrm{w})$ & Top wrong \\
\midrule
Gemma-27B & $+0.25$ $[0.13,0.38]$ & parity & 0.99 & 10, 12, 11 \\
Gemma-9B & $-0.30$ $[-0.42,-0.18]$ & fixed mid-range & 0.79 & 11, 12 \\
Llama-70B & $+0.05$ $[0.00,0.11]$ & under-count & 1.00 & 11, 10, 12 \\
Llama-8B & $+0.00$ $[0.00,0.00]$ & under-count & 1.00 & 10, 12, 9 \\
\bottomrule
\end{tabular}%
}
\caption{Within-family behavioral scale checks on the same $1{,}300$-prompt lattice (\textsc{common\_words} even$-$odd gap over $L\in\{10,\ldots,15\}$; bootstrap $95\%$ CIs).
$P(\mathrm{u}\mid\mathrm{w})=P(\hat{y}{<}y\mid$ wrong$)$.
Llama~8B matches Llama~70B's under-count class.
Gemma~9B's negative gap is a fixed mid-range collapse on this lattice, not a flipped 27B-style odd-to-even drop.}
\label{tab:scale-checks}
\end{table*}

\section{Full accuracy grid (Qwen~32B)}
\label{app:full-grid}

The main text reports pooled even$-$odd gaps.
Table~\ref{tab:full-accuracy} gives the underlying per-condition $\times$ per-length accuracy grid for the reference model Qwen~2.5--32B ($20$ lists per cell; $1{,}300$ prompts total).
Bold columns mark odd lengths $\{11,13,15\}$, where \textsc{common\_words} accuracy collapses.
Table~\ref{tab:full-signed} reports mean signed error (predicted $-$ true) on the same lattice: on \textsc{common\_words}, errors at $L{=}11$ tend toward $+1$ (say ``12'') and at $L{=}15$ toward $-1$ (say ``14''), while \textsc{ml\_mixed} under-counts severely.
Grids for the other six models are in Appendix~\ref{app:per-model-grid}.

\begin{table*}[!tbp]
\centering
\footnotesize
\setlength{\tabcolsep}{2.5pt}
\resizebox{\textwidth}{!}{%
\begin{tabular}{@{}l*{13}{c}@{}}
\toprule
condition & 3 & 4 & 5 & 6 & 7 & 8 & 9 & 10 & \textbf{11} & 12 & \textbf{13} & 14 & \textbf{15} \\
\midrule
\textsc{common\_words}        & 1.00 & 1.00 & 0.90 & 1.00 & 0.80 & 0.85 & 0.65 & 1.00 & \textbf{0.05} & 0.75 & \textbf{0.25} & 0.85 & \textbf{0.00} \\
\textsc{rare\_words}          & 1.00 & 1.00 & 0.95 & 0.65 & 0.80 & 0.80 & 0.80 & 1.00 & 0.30          & 0.95 & 0.50          & 0.85 & 0.15          \\
\textsc{ml\_single}           & 1.00 & 0.95 & 1.00 & 1.00 & 0.60 & 0.85 & 0.25 & 0.85 & 0.25          & 0.95 & 0.40          & 0.75 & 0.25          \\
\textsc{ml\_mixed}            & 1.00 & 1.00 & 0.85 & 0.60 & 0.60 & 0.70 & 0.00 & 0.55 & \textbf{0.00} & 0.50 & \textbf{0.05} & 0.05 & \textbf{0.00} \\
\textsc{random\_strings}      & 1.00 & 0.95 & 1.00 & 1.00 & 0.85 & 0.95 & 0.90 & 1.00 & 0.55          & 1.00 & 0.95          & 0.90 & 0.90          \\
\bottomrule
\end{tabular}%
}
\caption{Full accuracy grid for the reference model Qwen 2.5--32B, $L \in \{3,\ldots,15\}$ and the five surface conditions. $20$ sequences per cell; $1{,}300$ total. Bold marks odd lengths $\{11,13,15\}$.}
\label{tab:full-accuracy}
\end{table*}

\begin{table*}[!tbp]
\centering
\footnotesize
\setlength{\tabcolsep}{2.5pt}
\resizebox{\textwidth}{!}{%
\begin{tabular}{@{}l*{13}{c}@{}}
\toprule
 & 3 & 4 & 5 & 6 & 7 & 8 & 9 & 10 & \textbf{11} & 12 & \textbf{13} & 14 & \textbf{15} \\
\midrule
\textsc{common\_words}   & .00 & .00 & .00 & .00 & -.10 & -.05 & -.15 & .00 & \textbf{+.95} & +.25 & \textbf{+.15} & -.10 & \textbf{-.65} \\
\textsc{rare\_words}     & .00 & .00 & +.05 & +.35 & +.20 & +.10 & +.20 & .00 & -.20 & +.05 & -.20 & +.40 & .00 \\
\textsc{ml\_single}      & .00 & -.05 & .00 & .00 & -.20 & -.05 & -.35 & -.30 & +.25 & +.05 & -.50 & -.45 & -.45 \\
\textsc{ml\_mixed}       & .00 & .00 & -.05 & -.10 & -.40 & -.35 & -.75 & -.55 & -1.25 & -1.00 & -1.35 & -1.75 & -2.70 \\
\textsc{random\_strings} & .00 & -.05 & .00 & .00 & -.05 & -.05 & -.15 & .00 & -.45 & .00 & -.05 & -.10 & .00 \\
\bottomrule
\end{tabular}%
}
\caption{Mean signed error (predicted $-$ true) for Qwen 2.5--32B. On \textsc{common\_words}, $L{=}11$ errors are $+1$ on average (model says ``12''); $L{=}15$ errors are $-1$ on average (model says ``14''). \textsc{ml\_mixed} undercounts severely.}
\label{tab:full-signed}
\end{table*}

\section{Per-model accuracy grids}
\label{app:per-model-grid}

Tables~\ref{tab:grid-qwen14b}--\ref{tab:grid-llama70b} give the same per-condition $\times$ per-length accuracy grid for the six non-reference models on the shared $1{,}300$-prompt eval.
The Qwen~32B reference grid is Table~\ref{tab:full-accuracy} above.
Every cell is mean accuracy over $20$ sequences; seeds and lists are identical across models.

\begin{table*}[!tbp]
\centering
\footnotesize
\setlength{\tabcolsep}{2.5pt}
\resizebox{\textwidth}{!}{%
\begin{tabular}{@{}l*{13}{c}@{}}
\toprule
condition & 3 & 4 & 5 & 6 & 7 & 8 & 9 & 10 & \textbf{11} & 12 & \textbf{13} & 14 & \textbf{15} \\
\midrule
\textsc{common\_words}   & 0.95 & 0.95 & 0.80 & 0.95 & 0.50 & 0.75 & 0.50 & 0.75 & \textbf{0.20} & 0.95 & \textbf{0.05} & 0.15 & \textbf{0.00} \\
\textsc{rare\_words}     & 1.00 & 0.90 & 0.80 & 0.75 & 0.55 & 0.60 & 0.25 & 0.55 & 0.15          & 0.70 & 0.35          & 0.65 & 0.10          \\
\textsc{ml\_single}      & 1.00 & 0.85 & 0.80 & 0.90 & 0.65 & 0.85 & 0.20 & 0.75 & 0.15          & 0.70 & 0.55          & 0.30 & 0.20          \\
\textsc{ml\_mixed}       & 1.00 & 0.90 & 0.70 & 0.75 & 0.30 & 0.60 & 0.00 & 0.05 & \textbf{0.00} & 0.25 & \textbf{0.00} & 0.45 & \textbf{0.00} \\
\textsc{random\_strings} & 1.00 & 1.00 & 1.00 & 0.90 & 0.75 & 0.80 & 0.50 & 0.90 & 0.50          & 0.90 & 0.55          & 0.65 & 0.55          \\
\bottomrule
\end{tabular}%
}
\caption{Qwen 2.5--14B accuracy grid. Odd-length collapse at $\{11, 13, 15\}$ is sharp on every condition; \textsc{random\_strings} remains highest.}
\label{tab:grid-qwen14b}
\end{table*}

\begin{table*}[!tbp]
\centering
\footnotesize
\setlength{\tabcolsep}{2.5pt}
\resizebox{\textwidth}{!}{%
\begin{tabular}{@{}l*{13}{c}@{}}
\toprule
condition & 3 & 4 & 5 & 6 & 7 & 8 & 9 & 10 & \textbf{11} & 12 & \textbf{13} & 14 & \textbf{15} \\
\midrule
\textsc{common\_words}   & 1.00 & 1.00 & 1.00 & 1.00 & 0.95 & 0.95 & 0.85 & 0.80 & 0.95 & 0.60 & 0.85 & 0.60 & 0.65 \\
\textsc{rare\_words}     & 1.00 & 1.00 & 1.00 & 1.00 & 1.00 & 1.00 & 0.75 & 0.95 & 0.45 & 1.00 & 0.80 & 0.10 & 0.35 \\
\textsc{ml\_single}      & 1.00 & 1.00 & 0.95 & 1.00 & 1.00 & 0.85 & 0.50 & 0.80 & 0.55 & 0.70 & 0.70 & 0.50 & 0.40 \\
\textsc{ml\_mixed}       & 1.00 & 0.95 & 0.95 & 0.95 & 0.75 & 0.75 & 0.15 & 0.40 & 0.15 & 0.80 & 0.20 & 0.25 & 0.10 \\
\textsc{random\_strings} & 1.00 & 1.00 & 1.00 & 1.00 & 1.00 & 0.95 & 0.85 & 1.00 & 1.00 & 0.90 & 0.90 & 0.90 & 0.55 \\
\bottomrule
\end{tabular}%
}
\caption{Qwen 2.5--72B (4-bit nf4) accuracy grid. The odd-length collapse at $\{11, 13, 15\}$ is much weaker than at 14B/32B; the odd--even gap is partly preserved on \textsc{rare\_words} and \textsc{ml\_mixed} but absent on \textsc{common\_words}.}
\label{tab:grid-qwen72b}
\end{table*}

\begin{table*}[!tbp]
\centering
\footnotesize
\setlength{\tabcolsep}{2.5pt}
\resizebox{\textwidth}{!}{%
\begin{tabular}{@{}l*{13}{c}@{}}
\toprule
condition & 3 & 4 & 5 & 6 & 7 & 8 & 9 & 10 & \textbf{11} & 12 & \textbf{13} & 14 & \textbf{15} \\
\midrule
\textsc{common\_words}   & 1.00 & 1.00 & 0.90 & 0.55 & 0.25 & 0.15 & 0.15 & 0.75 & \textbf{0.10} & 0.10 & \textbf{0.00} & 0.00 & \textbf{0.00} \\
\textsc{rare\_words}     & 1.00 & 1.00 & 0.90 & 0.50 & 0.35 & 0.40 & 0.30 & 0.95 & 0.25          & 0.20 & 0.05          & 0.10 & 0.25          \\
\textsc{ml\_single}      & 1.00 & 1.00 & 0.65 & 0.20 & 0.15 & 0.30 & 0.40 & 0.55 & 0.05          & 0.15 & 0.00          & 0.00 & 0.00          \\
\textsc{ml\_mixed}       & 0.95 & 0.95 & 0.50 & 0.10 & 0.05 & 0.15 & 0.10 & 0.25 & \textbf{0.05} & 0.00 & \textbf{0.00} & 0.05 & \textbf{0.00} \\
\textsc{random\_strings} & 1.00 & 1.00 & 0.75 & 0.35 & 0.25 & 0.55 & 0.20 & 0.60 & 0.05          & 0.20 & 0.35          & 0.55 & 0.15          \\
\bottomrule
\end{tabular}%
}
\caption{Gemma 2--27B accuracy grid. Overall accuracy is lower across the board; the odd-length collapse is most visible on \textsc{common\_words}, \textsc{rare\_words}, and \textsc{ml\_mixed}.}
\label{tab:grid-gemma27b}
\end{table*}

\begin{table*}[!tbp]
\centering
\footnotesize
\setlength{\tabcolsep}{2.5pt}
\resizebox{\textwidth}{!}{%
\begin{tabular}{@{}l*{13}{c}@{}}
\toprule
condition & 3 & 4 & 5 & 6 & 7 & 8 & 9 & 10 & \textbf{11} & 12 & \textbf{13} & 14 & \textbf{15} \\
\midrule
\textsc{common\_words}   & 0.95 & 0.85 & 0.35 & 0.45 & 0.00 & 0.85 & 0.10 & 0.50 & \textbf{0.30} & 1.00 & \textbf{0.05} & 0.50 & \textbf{0.00} \\
\textsc{rare\_words}     & 1.00 & 1.00 & 0.80 & 0.90 & 0.30 & 0.95 & 0.60 & 0.50 & 0.50          & 0.85 & 0.10          & 0.65 & 0.00          \\
\textsc{ml\_single}      & 1.00 & 1.00 & 0.50 & 0.45 & 0.25 & 0.70 & 0.15 & 0.65 & 0.25          & 0.65 & 0.35          & 0.35 & 0.30          \\
\textsc{ml\_mixed}       & 0.90 & 0.70 & 0.20 & 0.15 & 0.00 & 0.15 & 0.00 & 0.10 & \textbf{0.00} & 0.40 & \textbf{0.00} & 0.25 & \textbf{0.00} \\
\textsc{random\_strings} & 1.00 & 0.70 & 0.55 & 0.75 & 0.70 & 1.00 & 0.25 & 0.95 & 0.60          & 0.85 & 0.40          & 0.65 & 0.65          \\
\bottomrule
\end{tabular}%
}
\caption{DeepSeek-R1-Distill-Qwen-32B accuracy grid (reasoning enabled). The odd-length collapse at $\{11, 13, 15\}$ is preserved through R1 distillation; even--odd gap (0.39) is essentially identical to its base Qwen 2.5--32B (0.40).}
\label{tab:grid-r1qwen32b}
\end{table*}

\begin{table*}[!tbp]
\centering
\footnotesize
\setlength{\tabcolsep}{2.5pt}
\resizebox{\textwidth}{!}{%
\begin{tabular}{@{}l*{13}{c}@{}}
\toprule
condition & 3 & 4 & 5 & 6 & 7 & 8 & 9 & 10 & \textbf{11} & 12 & \textbf{13} & 14 & \textbf{15} \\
\midrule
\textsc{common\_words}   & 1.00 & 0.55 & 1.00 & 0.90 & 0.30 & 0.00 & 0.10 & 0.00 & 0.00 & 0.00 & 0.00 & 0.00 & 0.00 \\
\textsc{rare\_words}     & 0.95 & 0.35 & 0.85 & 0.30 & 0.50 & 0.15 & 0.00 & 0.00 & 0.35 & 0.10 & 0.00 & 0.50 & 0.00 \\
\textsc{ml\_single}      & 0.95 & 0.45 & 0.80 & 0.65 & 0.75 & 0.25 & 0.60 & 0.20 & 0.15 & 0.10 & 0.00 & 0.40 & 0.15 \\
\textsc{ml\_mixed}       & 0.70 & 0.05 & 0.55 & 0.55 & 0.65 & 0.50 & 0.55 & 0.15 & 0.65 & 0.10 & 0.00 & 0.80 & 0.10 \\
\textsc{random\_strings} & 0.85 & 0.75 & 0.55 & 0.60 & 0.40 & 0.45 & 0.20 & 0.25 & 0.85 & 0.35 & 0.00 & 0.65 & 0.00 \\
\bottomrule
\end{tabular}%
}
\caption{OLMo 2--32B accuracy grid. Overall accuracy is lower than the Qwen/Gemma row even at short lengths, but there is no odd-length collapse: odd lengths are no worse than adjacent even lengths on most conditions. This is the no-collapse baseline that the cross-family argument in \S\ref{sec:results:cross-model} rests on.}
\label{tab:grid-olmo32b}
\end{table*}

\begin{table*}[!tbp]
\centering
\footnotesize
\setlength{\tabcolsep}{2.5pt}
\resizebox{\textwidth}{!}{%
\begin{tabular}{@{}l*{13}{c}@{}}
\toprule
condition & 3 & 4 & 5 & 6 & 7 & 8 & 9 & 10 & \textbf{11} & 12 & \textbf{13} & 14 & \textbf{15} \\
\midrule
\textsc{common\_words}   & 1.00 & 1.00 & 0.90 & 1.00 & 0.75 & 0.75 & 0.10 & 0.15 & \textbf{0.00} & 0.00 & \textbf{0.00} & 0.00 & \textbf{0.00} \\
\textsc{rare\_words}     & 1.00 & 1.00 & 1.00 & 0.95 & 0.80 & 0.95 & 0.55 & 0.30 & \textbf{0.95} & 0.55 & \textbf{0.50} & 0.00 & \textbf{0.70} \\
\textsc{ml\_single}      & 0.90 & 0.90 & 0.65 & 0.80 & 0.65 & 0.45 & 0.55 & 0.35 & \textbf{0.80} & 0.05 & \textbf{0.50} & 0.10 & \textbf{0.30} \\
\textsc{ml\_mixed}       & 1.00 & 0.95 & 0.95 & 1.00 & 0.75 & 0.35 & 0.70 & 0.45 & \textbf{0.95} & 0.05 & \textbf{0.15} & 0.05 & \textbf{0.40} \\
\textsc{random\_strings} & 0.95 & 1.00 & 0.75 & 0.90 & 0.45 & 0.65 & 0.35 & 0.30 & \textbf{0.75} & 0.25 & \textbf{0.40} & 0.00 & \textbf{0.30} \\
\bottomrule
\end{tabular}%
}
\caption{Llama 3.1--70B (4-bit nf4) accuracy grid. On \textsc{common\_words}, the model is accurate at short lengths and then collapses above $L\!=\!8$ to a small-integer under-counting attractor rather than to the Qwen/Gemma odd-length collapse.}
\label{tab:grid-llama70b}
\end{table*}

\section{Linear count probe: extrapolation falsifier}
\label{app:probe-extrap}

\begin{figure}[t]
    \centering
    \includegraphics[width=0.95\linewidth]{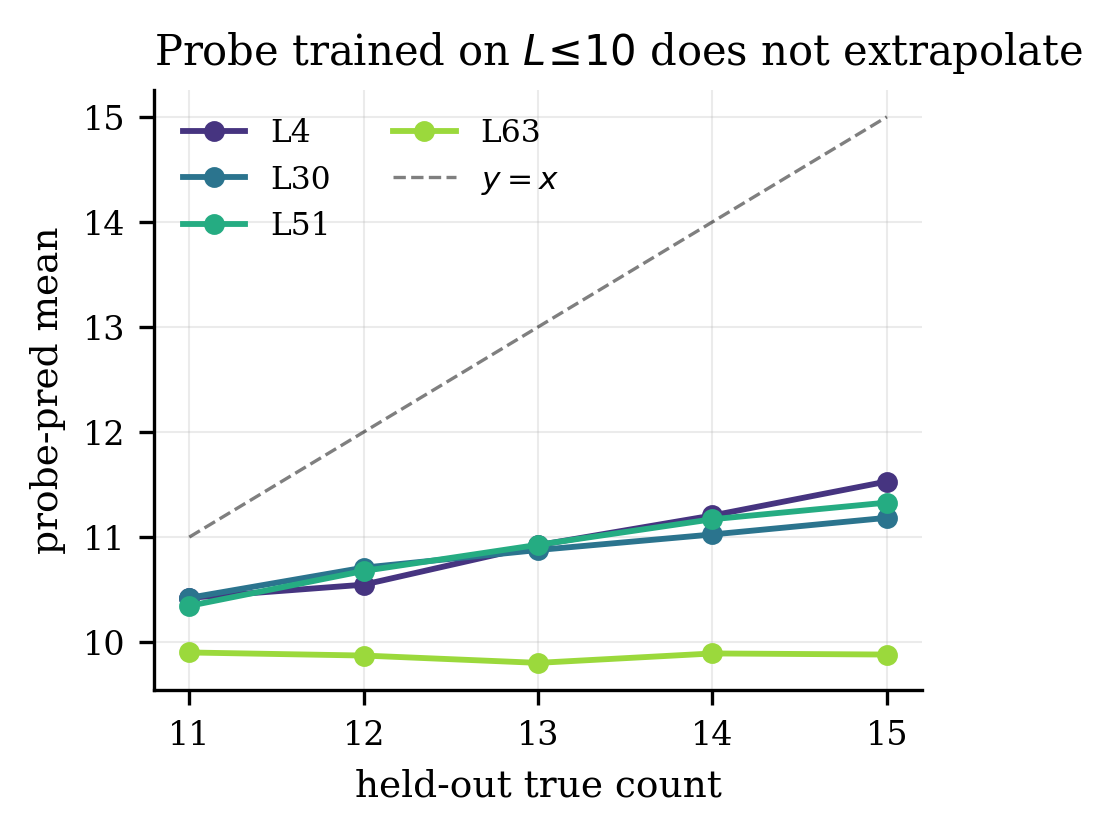}
    \caption{Length-extrapolation falsifier. Mean probe-predicted count, by held-out true count, for a probe trained on $L \le 10$. At every layer (only four shown), the predicted mean collapses to the train-set ceiling ($\sim\!9.9$) for every held-out count; round-accuracy is $0.00$. The dashed line is $y = x$, the prediction a single global cardinality axis would produce.}
    \label{fig:probe-extrap}
\end{figure}

Figure~\ref{fig:probe-extrap} shows the length-extrapolation probe of \S\ref{sec:results:probe} in detail. The probe is a ridge regressor ($\alpha = 10$) fit on the 320 \textsc{common\_words} sequences with $L \in \{3,\ldots,10\}$ and evaluated on the 200 held-out sequences with $L \in \{11,\ldots,15\}$. We plot the mean predicted count, by held-out true count, at four representative layers ($\ell_{4}, \ell_{30}, \ell_{51}, \ell_{63}$). At every layer the predicted mean for every held-out count flattens to roughly the largest trained count ($\sim\!9.9$); round-accuracy is $0.00$ at every test length and every layer. By contrast, the leave-one-count-out probe of \S\ref{sec:results:probe}, which trains on every count except a single held-out one, achieves round-accuracy $\ge 0.70$ on the held-out count for $L^\star \in \{11,12,13,14\}$. Together these falsify the claim that the residual contains a single linear ``cardinality direction'' that can be extended indefinitely, while preserving the claim that the count is approximately linearly decodable within the trained range.

\section{Linear count probe: per-length detail}
\label{app:probe-per-length}

Table~\ref{tab:probe-per-length} reports the per-length mean predicted count and round-accuracy of the layer-$63$ ridge probe ($\alpha = 1$, $5$-fold CV) on the $520$-sequence \textsc{common\_words} cell, alongside the reference Qwen-32B model's own greedy accuracy on the same prompts. The probe is at or above $0.85$ round-accuracy at every length except $L{=}12$ ($0.75$); the model collapses to $0.05$/$0.25$/$0.00$ at odd lengths $\{11,13,15\}$. The gap at $L{=}11$ is $0.85 - 0.05 = 0.80$.

\begin{table}[h]
\centering\small
\setlength{\tabcolsep}{4pt}
\begin{tabular}{ccccc}
\toprule
$L$ & probe pred mean & probe round-acc & model acc \\
\midrule
3  & $3.09$  & $0.95$ & $1.00$ \\
4  & $4.01$  & $1.00$ & $1.00$ \\
5  & $5.01$  & $1.00$ & $0.90$ \\
6  & $6.01$  & $0.95$ & $1.00$ \\
7  & $7.02$  & $0.98$ & $0.80$ \\
8  & $7.98$  & $0.98$ & $0.85$ \\
9  & $9.02$  & $0.98$ & $0.65$ \\
10 & $10.04$ & $0.93$ & $1.00$ \\
\textbf{11} & $11.00$ & $\mathbf{0.85}$ & $\mathbf{0.05}$ \\
12 & $11.98$ & $0.75$ & $0.75$ \\
\textbf{13} & $13.06$ & $\mathbf{0.85}$ & $\mathbf{0.25}$ \\
14 & $13.94$ & $0.88$ & $0.85$ \\
\textbf{15} & $14.86$ & $\mathbf{0.93}$ & $\mathbf{0.00}$ \\
\bottomrule
\end{tabular}
\caption{Per-length probe vs.~model accuracy at the answer-position residual (layer $63$, Qwen-32B). The probe column is the held-out (5-fold CV) ridge prediction; the model column is the model's own greedy emit on the same prompts. Bold rows are the odd-length collapse. Overall: $r = 0.997$, $\text{MAE} = 0.20$, round-acc $0.923$.}
\label{tab:probe-per-length}
\end{table}

\section{Causal intervention protocols}
\label{app:interventions}

\paragraph{Block decomposition.}
For a single failure example $(x, y_{\text{correct}}, y_{\text{wrong}})$ with $x$ the answer-position residual entering block $\ell$, let $a_\ell$ and $m_\ell$ denote the attention and MLP outputs of that block, so that the block update is $\Delta_\ell = a_\ell + m_\ell$. We define the (correct $-$ wrong) digit-margin direction in residual space as $\Delta_u = W_U[:, y_{\text{correct}}] - W_U[:, y_{\text{wrong}}]$, where $W_U$ is the unembedding matrix. Projections are $\langle a_\ell, \Delta_u\rangle$ and $\langle m_\ell, \Delta_u\rangle$; norms are $\ell_2$ norms in the residual basis. Layer $\ell{=}62$ was selected as the late layer with the largest negative MLP projection on $\Delta_u$ over the eight near-miss failures.

\paragraph{Per-failure layer search (\S\ref{sec:results:block-decomp}).}
For each of the eight near-miss failures we replace $\text{mlp}_\ell(x)$ with one of $\{0,\, -\tfrac{1}{2}\text{mlp}_\ell(x)\}$ ("zero" / "reverse-half") at one layer $\ell$ at a time, sweeping $\ell \in \{34, 41, 46, 47, 49, 50, 51, 52, 54, 57\text{--}63\}$, and re-running the forward pass. A failure is counted as "fixed at $\ell$" when both the argmax digit becomes the correct digit and the (correct $-$ wrong) margin becomes positive. Layer $52$ \texttt{reverse\_half} is the single $(\ell, \text{intervention})$ pair that fixes the largest number ($6/8$) of these failures.

\paragraph{Split-half layer search (selection-corrected).}
For the Qwen-32B population estimate, the same \texttt{reverse\_half} intervention is applied at each of $16$ late MLP layers ($\ell \in \{34, 41, 46, 47, 49, 50, 51, 52, 54, 57,\ldots,63\}$) to every one of the $102$ \textsc{common\_words} odd-length failures ($L\in\{11,13,15\}$, model wrong), one $(\ell, \text{failure})$ cell at a time. For each of $20$ random halvings of the failure set, the layer with the highest fix rate on the selection half is evaluated on the disjoint evaluation half. Reported statistics are the distribution of selected layers and the mean $\pm$ SD of the evaluation-half fix rate.
The Qwen-14B estimate uses the identical protocol on all $103$ probe-cell odd-length failures over the last $16$ MLP layers ($\ell \in \{32,\ldots,47\}$ for the $48$-layer stack), yielding held-out $0.25\pm0.05$ with $\ell_{33}$ selected in $16/20$ splits.
Gemma-27B uses the same full-failure protocol on all $112$ probe-cell odd-length failures over the last $16$ MLP layers of the $46$-layer stack ($\ell \in \{30,\ldots,45\}$), yielding held-out $0.055\pm0.022$; an earlier $n{=}8$ vignette that suggested ${\sim}0.69$ was selection bias and is not used as an estimate.

\paragraph{Filler-padding capacity check (not a dilution refutation).}
Softmax-dilution accounts concern resolution over the $N$ counted items \citep{velivckovic2025softmax}. Prepending neutral filler at fixed $N$ changes context length, not the number of competing count-relevant targets, so this is only an answer-position capacity check. For $N\in\{5,13\}$ and $n{=}40$ fresh \textsc{common\_words} lists per cell on Qwen~32B, the user message is either the standard prompt or the same prompt preceded by $40$ repetitions of a neutral filler sentence (${\sim}760$ tokens), inside a single user turn (${\sim}55\to{\sim}815$ tokens). Greedy decoding: $N{=}5$ accuracy moves from $1.00$ to $0.90$ and $N{=}13$ from $0.05$ to $0.00$. Long-context padding is not the odd-length bottleneck; we do not treat this result as evidence against softmax dilution.

\paragraph{Multi-word item control.}
Lists of $N$ items in which exactly one item is a familiar two-word phrase (e.g.\ ``new york''), so item count $N$ and whitespace-chunk count $N{+}1$ differ. $n{=}40$ lists per $N$, $N\in\{10,\ldots,14\}$; we report the fraction of answers equal to $N$ (item counting) vs.\ $N{+}1$ (chunk counting).

\paragraph{Class-mean steering.}
From the cached probe-cell residuals we compute per-count class means $\mu_c$ at layer $\ell\in\{49, 52\}$. For each odd-length failure with true count $N$ and wrong emission $\hat{y}$, a forward hook adds $\beta\,(\mu_N - \mu_{\hat{y}})$ (with $\beta\in\{2,4,8\}$) to the last-position residual at layer $\ell$ on every forward pass during greedy decoding. Controls: a norm-matched fixed random direction per example, a same-day unsteered baseline, and a $40$-sequence model-correct sample steered away from its nearest attractor to measure collateral breakage. The probe-hyperplane projection variant (replacing the residual's probe read-out value with the target count) is reported as a null.

\paragraph{Global scale sweep.}
A forward hook multiplies the chosen layer's MLP \emph{module output} by a scalar $s$ before the residual add, i.e.\ $\mathrm{mlp}_\ell(x)\leftarrow s\cdot\mathrm{mlp}_\ell(x)$, for $s \in \{0.00,\, 0.25,\, 0.50,\, 0.75,\, 1.00,\, 1.25\}$.
Thus $s{=}1$ is the identity (unsteered baseline) and $s{=}0$ zeros that MLP contribution; values $s\neq 1$ are the intervention.
The continuous trade-off quoted in \S\ref{sec:results:block-decomp} (even $0.755{\to}0.865$, odd $0.653{\to}0.581$) uses the full main eval at the discovery layer $\ell{=}52$.
Figure~\ref{fig:cross-model-causal} (bottom) shows the same qualitative even/odd trade-off; axis labels may name a nearby late layer chosen by a later candidate-layer search (e.g.\ $\ell{=}58$ on a cliff subsample). Both are late-MLP magnitude sweeps, not a claim that a single index is uniquely causal.
``Even-target'' / ``odd-target'' accuracy average correctness over even / odd true counts in $\{3,\ldots,15\}$.

\paragraph{Layer-index map (Qwen-32B).}
Different analyses pick different late layers, and we do not equate them: block-decomposition margin projection on $n{=}8$ near-misses highlights $\ell{=}62$; the same vignette's best single-layer \texttt{reverse\_half} is $\ell{=}52$ ($6/8$); selection-corrected full-failure split-half most often selects $\ell{=}49$; the continuous magnitude sweep above uses $\ell{=}52$ (full eval) or a nearby late candidate in the figure panel.
Top panels of Figure~\ref{fig:cross-model-causal} plot the late-layer candidate subset used in each model's search, not every transformer layer.

\paragraph{SwiGLU feature decomposition (\S\ref{sec:results:feature}).}
A SwiGLU MLP block computes $\text{mlp}(h) = W_{\text{down}}\bigl(\text{SiLU}(W_{\text{gate}}h)\odot W_{\text{up}}h\bigr) = W_{\text{down}}\,z$, where $z \in \mathbb{R}^{d_{\text{ff}}}$ is the SwiGLU hidden vector and $d_{\text{ff}}{=}27{,}648$ for Qwen-32B. For the failure example in \S\ref{sec:results:feature}, we computed the per-index contribution to the (correct $-$ wrong) margin, $c_k = z_k \cdot \langle W_{\text{down}}[:, k],\, \Delta_u\rangle$, and selected $k^{\star} = \arg\min_k c_k$. Single-feature patches replace $z_{k^{\star}}$ with one of $\{0,\, \tfrac{1}{2}z_{k^{\star}},\, -\tfrac{1}{2}z_{k^{\star}},\, -z_{k^{\star}}\}$ while leaving every other index of $z$ untouched, and the forward pass is re-run from layer $52$ onward.

\section{Tokenization control across models}
\label{app:tokenization-control}

A pure tokenization-load account predicts that lists whose items fragment into more subword tokens should be harder to count.
Table~\ref{tab:tokenization-control} in the main text tests that prediction: every model is more accurate on \textsc{random\_strings} than on \textsc{common\_words} despite roughly $4\times$ more tokens at $L{=}15$ (\S\ref{sec:results:cliff}).

\section{Error-mode separability (not model fingerprinting)}
\label{app:error-fingerprint}

We ask how well counting answers on this fixed benchmark separate models. Features are answer integers, signed errors, length, and condition; splits are by prompt text so no prompt appears in both train and test. Chance is $1/7$ for model identity and $1/4$ for family (Qwen, Gemma, OLMo, Llama; R1-Distill is grouped with Qwen).

\begin{table}[t]
\centering
\small
\setlength{\tabcolsep}{3.5pt}
\begin{tabular}{lrrl}
\toprule
Setting & Acc & Chance & Note \\
\midrule
one answer $\to$ model & 0.27 & 0.14 & \\
one answer $\to$ family & 0.62 & 0.25 & \\
one wrong answer $\to$ model & 0.39 & 0.14 & \\
bag of 40 wrong answers $\to$ model & 0.93 & 0.14 & \\
bag of wrong answers $\to$ family & 1.00 & 0.25 & \\
\bottomrule
\end{tabular}
\caption{Error-mode separability on the shared benchmark. Single answers are weak identifiers; bags of wrong answers separate models well in this closed seven-model set. Family identity separates almost perfectly at the bag level. This is descriptive structure, not a claim that counting errors fingerprint models in the wild.}
\label{tab:error-fingerprint}
\end{table}

\begin{table*}[t]
\centering
\small
\setlength{\tabcolsep}{4pt}
\begin{tabular}{llrlr}
\toprule
Model & Family & Err rate & Top wrong emissions & Even share \\
\midrule
Gemma 27B & gemma & 0.62 & 10:0.26, 12:0.13, 11:0.09, 5:0.08, 8:0.08 & 0.64 \\
Llama 70B & llama & 0.43 & 11:0.30, 13:0.13, 12:0.12, 10:0.09, 9:0.07 & 0.35 \\
OLMo 32B & olmo & 0.62 & 11:0.25, 14:0.17, 9:0.11, 7:0.11, 8:0.09 & 0.40 \\
Qwen 14B & qwen & 0.42 & 12:0.29, 14:0.13, 8:0.13, 10:0.11, 6:0.10 & 0.78 \\
Qwen 32B & qwen & 0.29 & 12:0.27, 10:0.18, 14:0.13, 8:0.12, 7:0.07 & 0.82 \\
Qwen 72B & qwen & 0.21 & 12:0.25, 13:0.23, 10:0.14, 14:0.11, 11:0.09 & 0.61 \\
R1-Distill & qwen & 0.50 & 12:0.19, 8:0.17, 6:0.13, 4:0.11, 14:0.10 & 0.82 \\
\bottomrule
\end{tabular}
\caption{Wrong-emission modes on the $1{,}300$-prompt eval. \emph{Even share} is the fraction of wrong emissions that are even integers. Qwen-family models are even-heavy; among OLMo's wrong answers, mass peaks at ``$11$''; Llama is odd-heavy / under-counting.}
\label{tab:wrong-emit-modes}
\end{table*}

Single answers are weak identifiers (model accuracy $0.27$--$0.39$). Bags of wrong answers separate the closed seven-model set well ($0.93$), and family identity is essentially perfect at the bag level. Residual confusions are mostly inside the Qwen family. We report this as descriptive separability on this benchmark, not as a fingerprinting method for arbitrary models or prompts.

\section{Additional Visual Diagnostics}
\label{app:visual-diagnostics}

These plots support claims already stated in the main text; they are optional reading once the grids and intervention protocols above are clear.

\begin{figure*}[!tbp]
    \centering
    \includegraphics[width=0.92\textwidth]{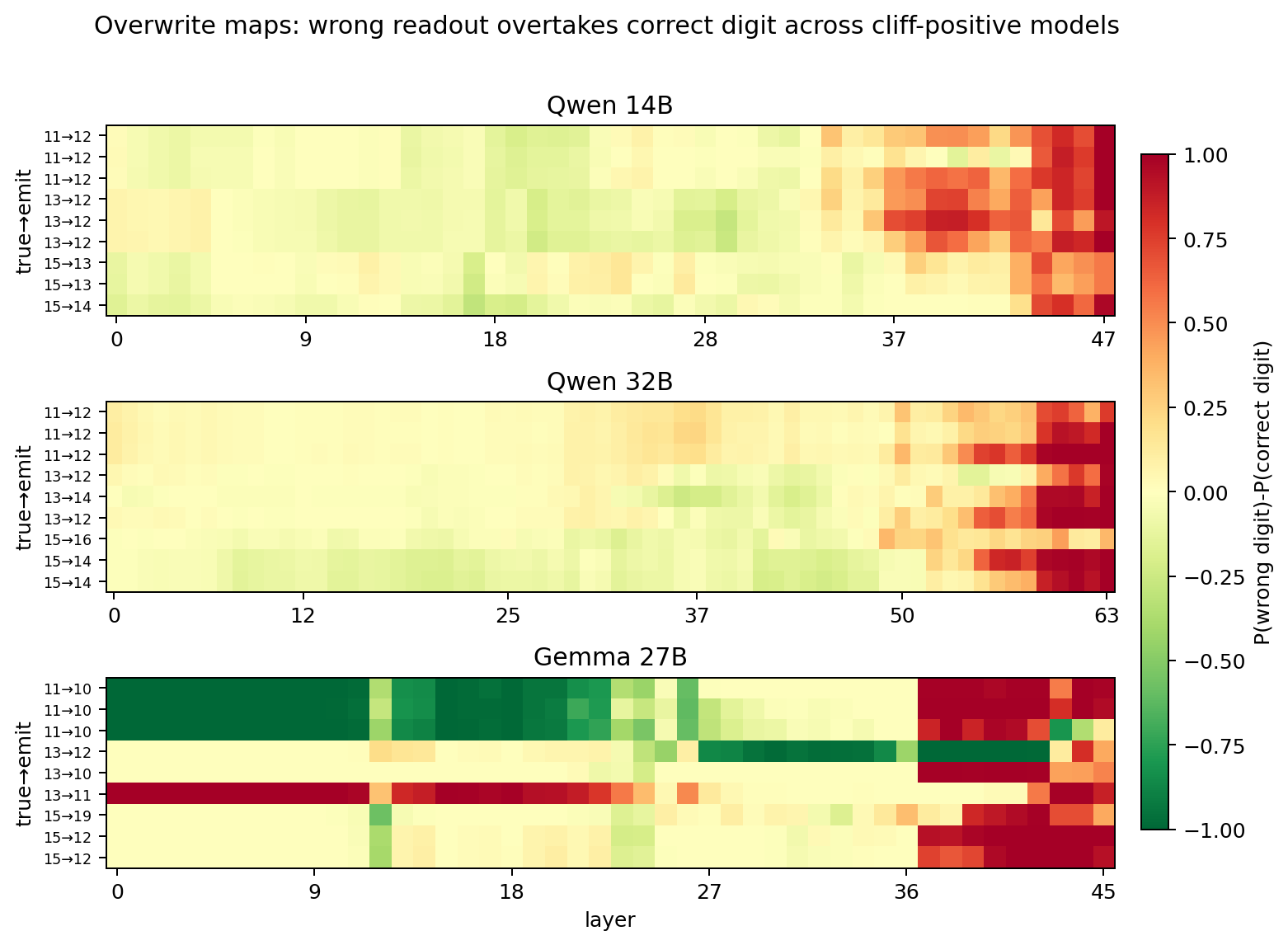}
    \caption{Cross-model overwrite maps for sampled odd-length failures. Each row is one prompt; color shows $P(\text{model's wrong units digit}) - P(\text{correct units digit})$ under a layerwise logit-lens readout. Green regions indicate layers where the correct digit is favored; red regions indicate layers where the eventual wrong digit dominates. Qwen14, Qwen32, and Gemma27B all show a mid-to-late transition toward the wrong digit, a \emph{read-only} diagnostic of late failure. Causal \texttt{reverse\_half} confirms a usable late-MLP lever only in Qwen (\S\ref{sec:results:block-decomp}).}
    \label{fig:cross-model-overwrite-map}
\end{figure*}

\begin{figure*}[!tbp]
    \centering
    \includegraphics[width=0.95\textwidth]{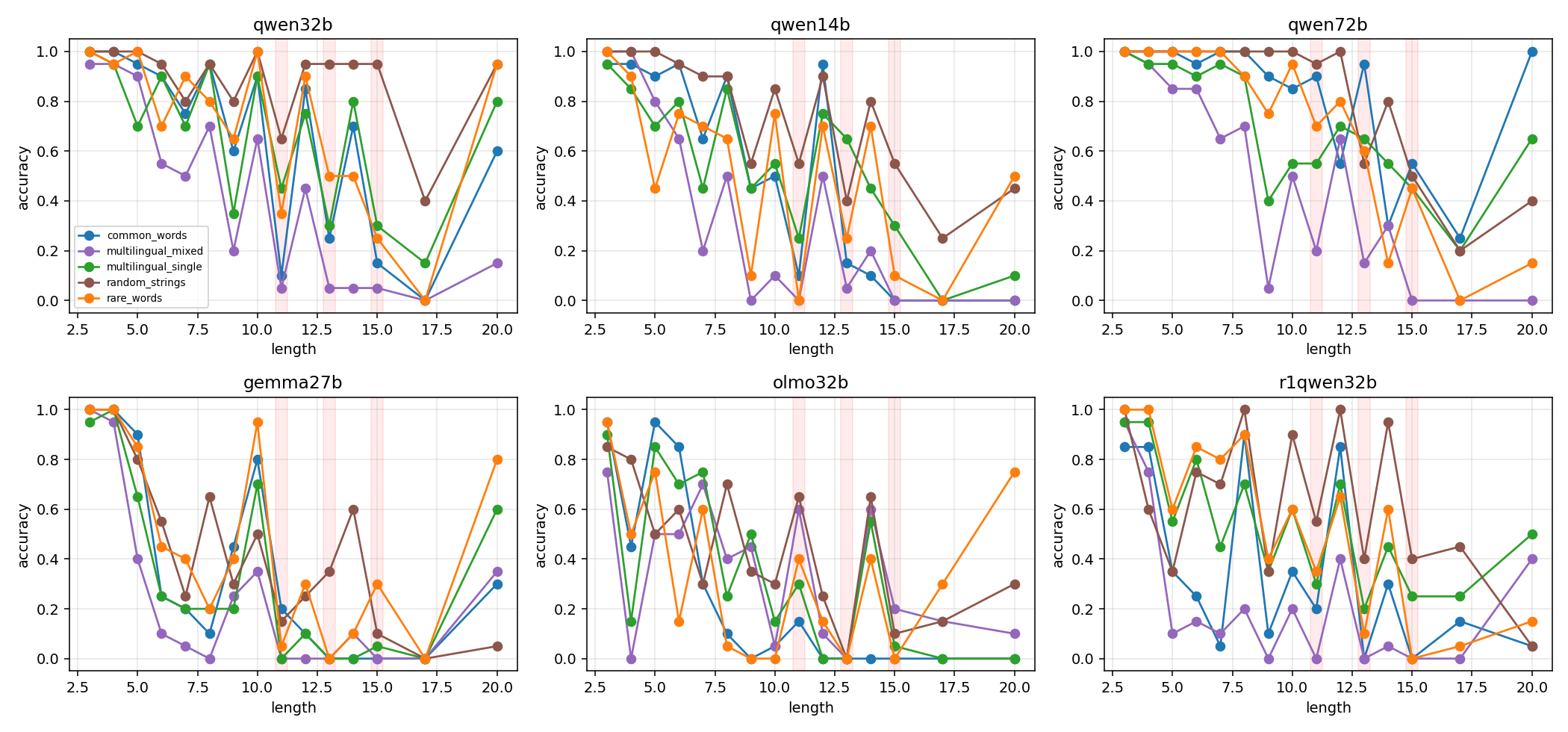}
    \caption{Accuracy vs.\ list length by surface condition (odd lengths $\{11,13,15\}$ shaded). Read for the cross-panel \emph{shape} split (parity oscillation vs.\ flat vs.\ under-count), not per-condition zig-zags; main-text summary is Table~\ref{tab:even-odd-gap} and Fig.~\ref{fig:main-overview}.}
    \label{fig:cross-model-cliff}
\end{figure*}

\begin{figure*}[!tbp]
    \centering
    \includegraphics[width=0.90\textwidth]{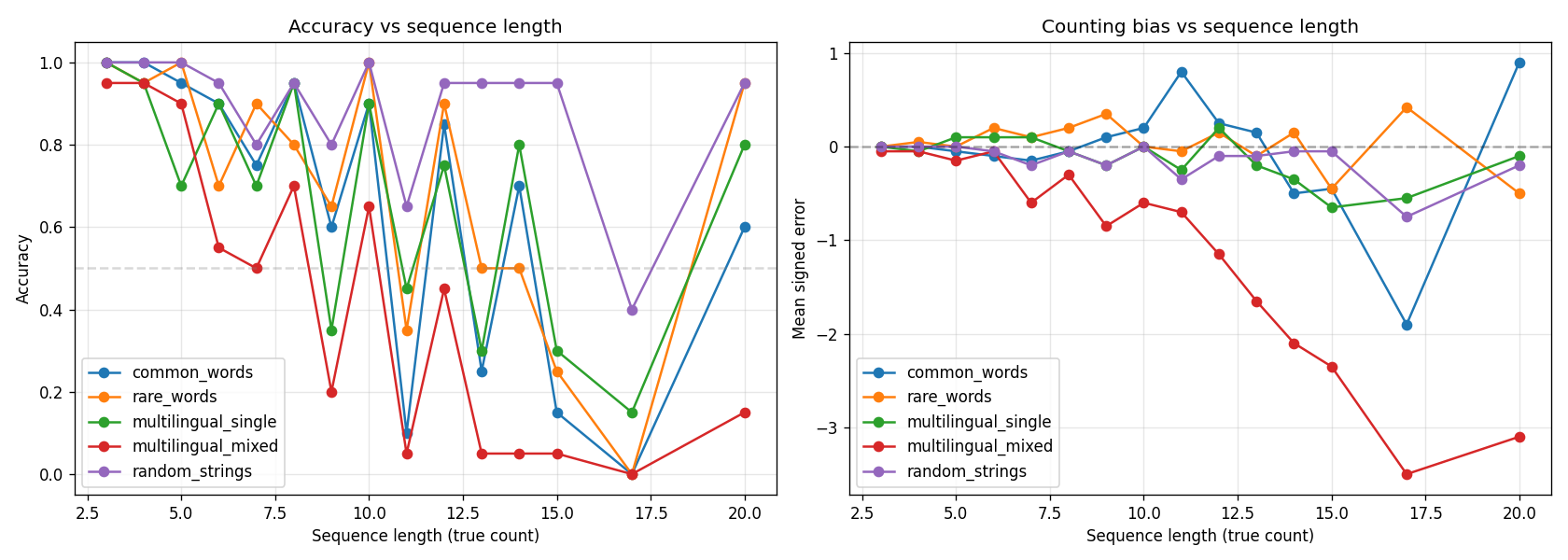}
    \caption{Reference-model odd-length collapse and signed-error diagnostics for Qwen 2.5--32B. The odd-length failures are directional rather than diffuse: $L=11$ tends upward toward ``12'', while $L=15$ tends downward toward ``14''.}
    \label{fig:app-qwen-cliff}
\end{figure*}

\begin{figure*}[!tbp]
    \centering
    \includegraphics[width=0.85\textwidth]{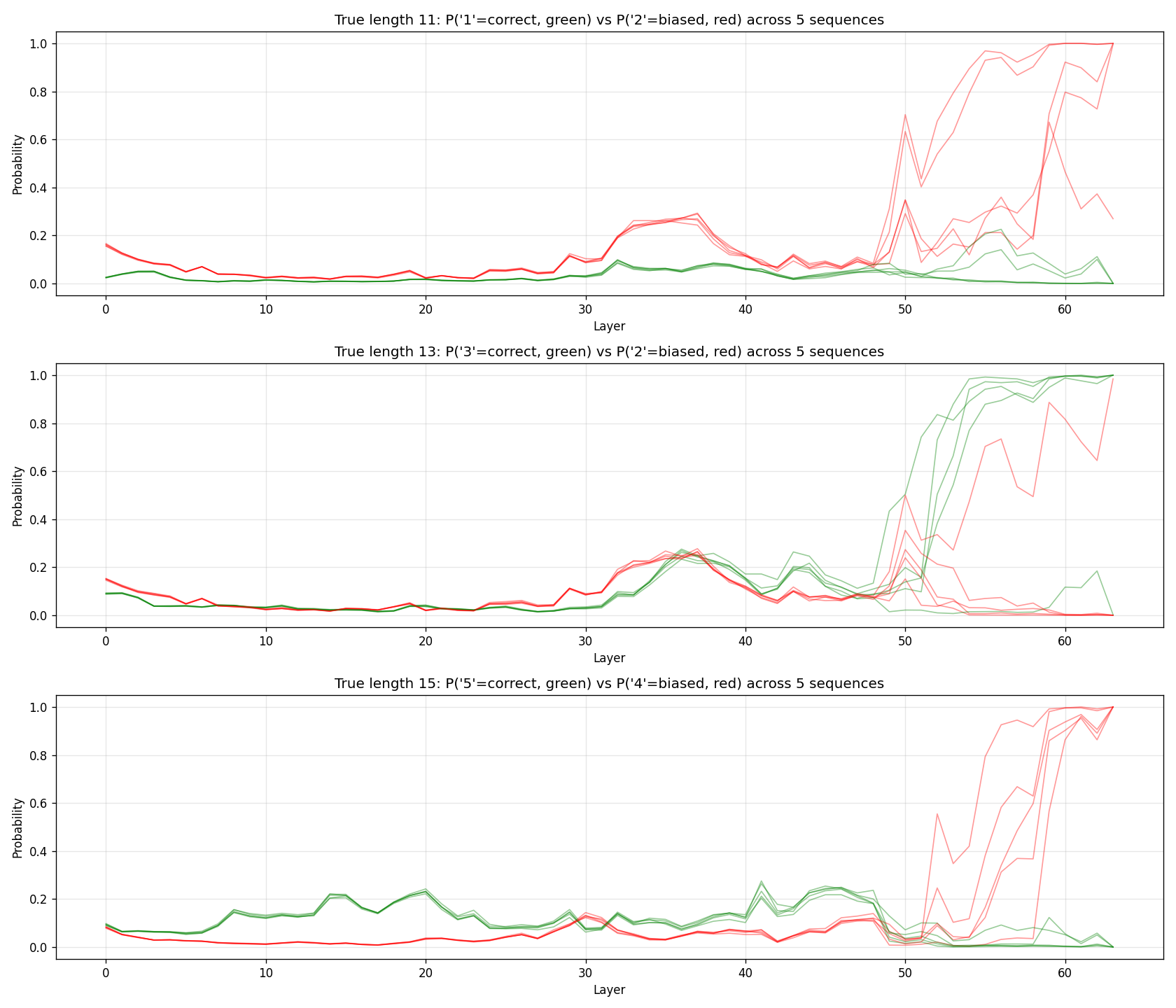}
    \caption{Qwen 2.5--32B logit-lens trajectories. Correct-digit mass appears in middle layers and is suppressed in late layers, consistent with a late overwrite rather than an absent count signal.}
    \label{fig:app-logit-lens}
\end{figure*}

\begin{figure*}[!tbp]
    \centering
    \includegraphics[width=0.96\textwidth]{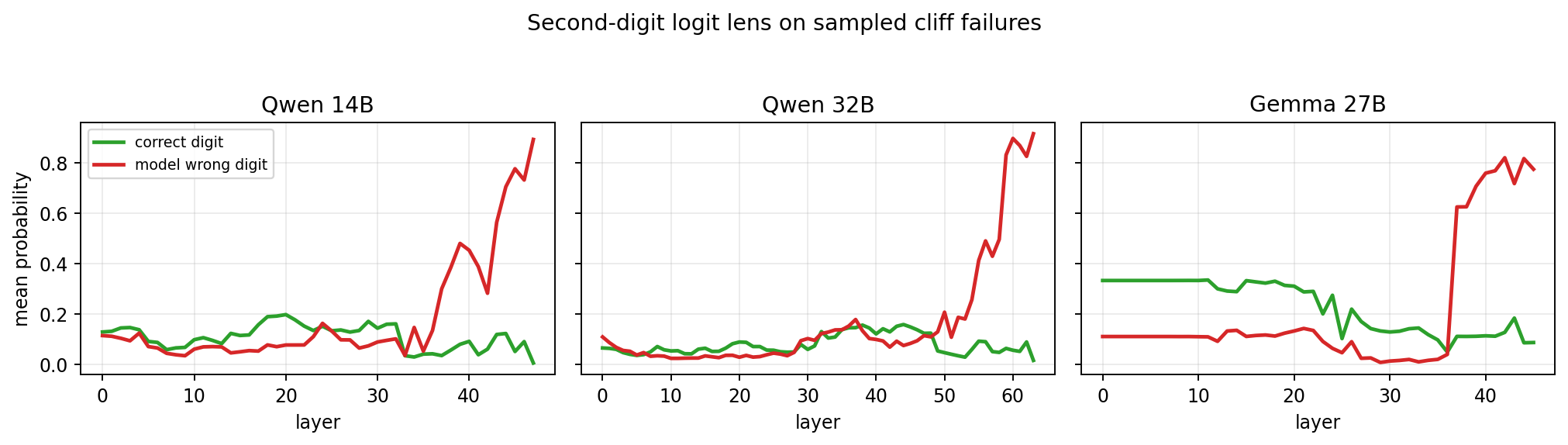}
    \caption{Cross-model second-digit logit lens on odd-length failures. For each of Qwen 14B, Qwen 32B, and Gemma 2 27B, we force the leading ``$1$'' on length-$\{11,13,15\}$ failures and read off the per-layer probability of the correct vs.\ wrong units digit. In every model the correct digit is favored at intermediate depth and the wrong digit dominates only late, with mean final $P(\text{wrong})\in[0.77, 0.92]$. Read-only evidence of late failure, not by itself a claim that the same MLP lever is causal in every family (Fig.~\ref{fig:cross-model-causal}).}
    \label{fig:second-digit-logit-lens}
\end{figure*}

\begin{figure}[!tbp]
    \centering
    \includegraphics[width=0.85\linewidth]{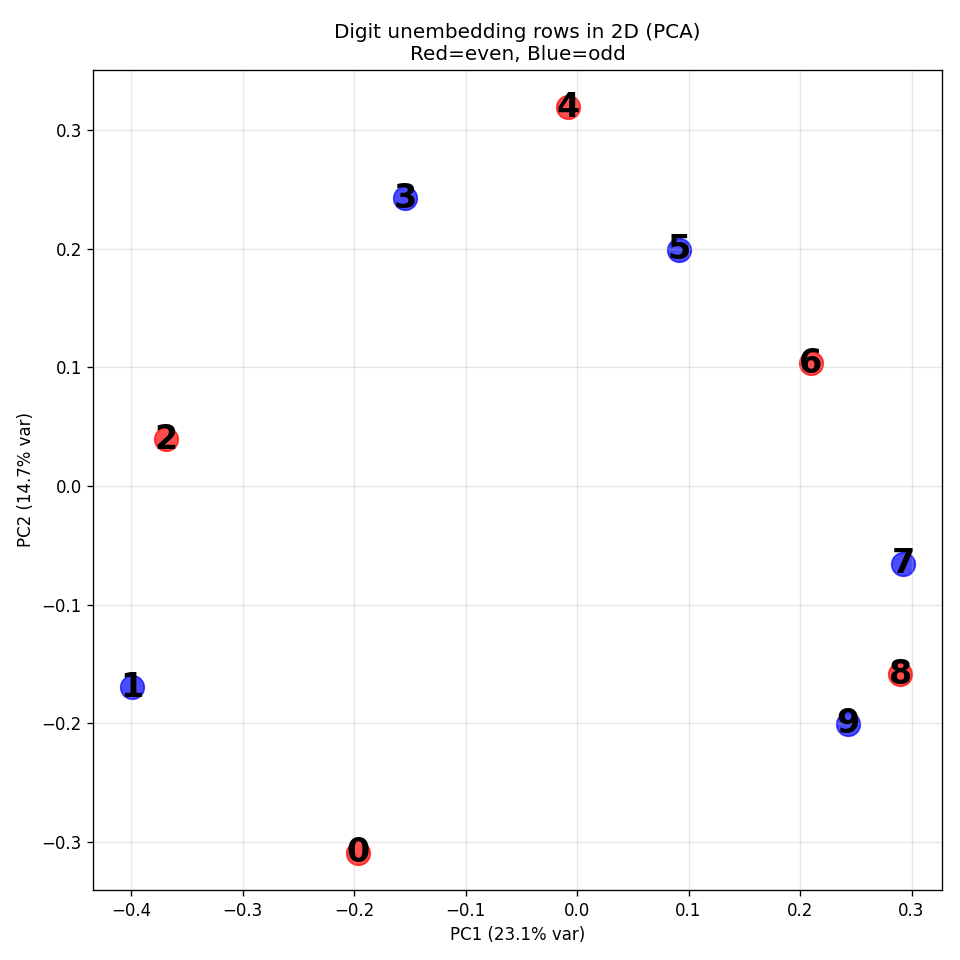}
    \caption{Digit-unembedding PCA for Qwen 2.5--32B. Digit geometry is ordinal; even/odd is not cleanly separated, supporting the claim that the parity pattern is not a simple head-side parity direction.}
    \label{fig:app-digit-pca}
\end{figure}

\begin{figure*}[!tbp]
    \centering
    \includegraphics[width=0.92\textwidth]{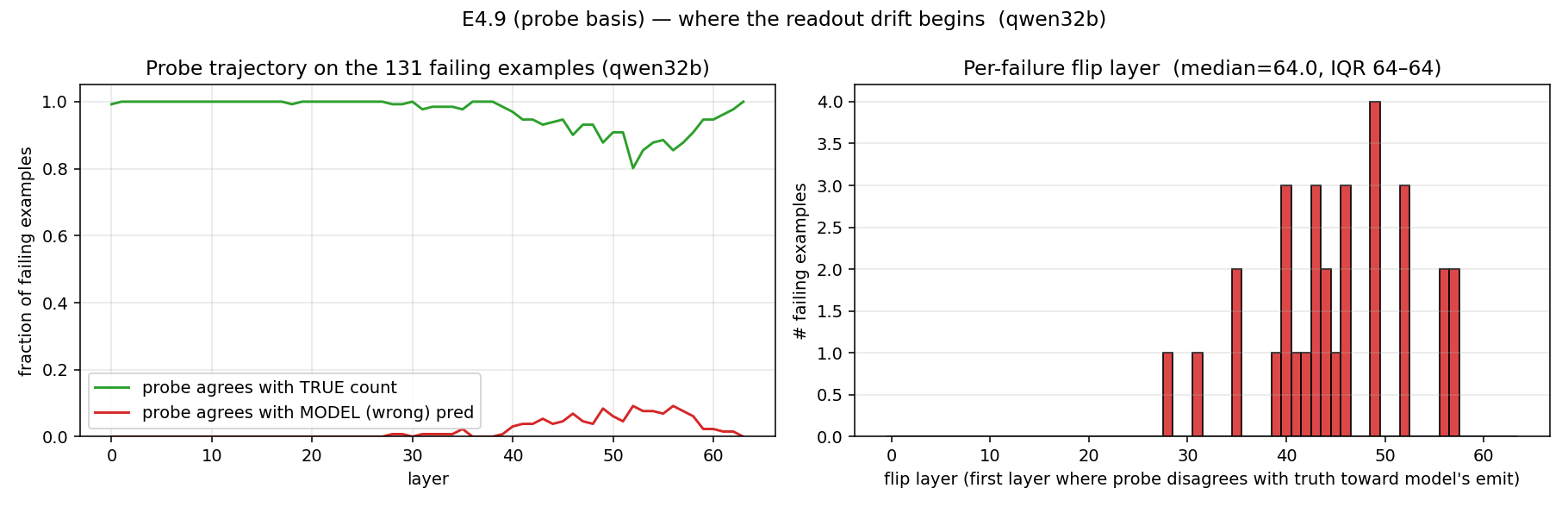}
    \caption{Probe-basis per-failure flip-layer analysis for Qwen 2.5--32B. The plot tracks when the probe agrees with the true count versus the model's wrong emitted count on probe-right/model-wrong examples. It is a probe-basis diagnostic, not the model's own logit-lens basis, and complements the logit-lens trajectories in Figure~\ref{fig:app-logit-lens}.}
    \label{fig:app-probe-flip-layer}
\end{figure*}

\begin{figure*}[!tbp]
    \centering
    \includegraphics[width=0.92\textwidth]{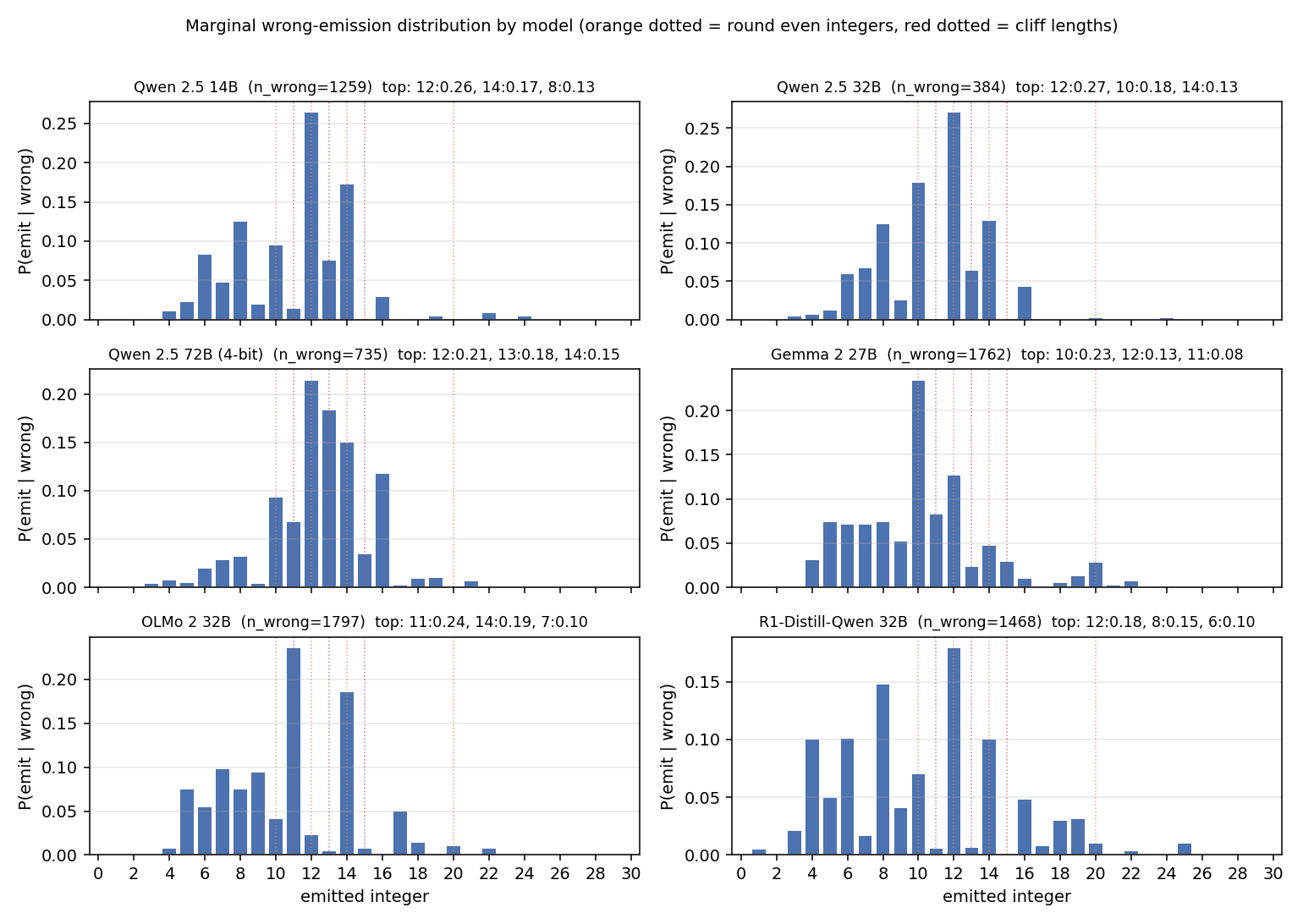}
    \caption{Model-output prior proxy: marginal wrong-emission distributions by model. Each family concentrates wrong answers on a small set of integers, but the attractor set differs across families.}
    \label{fig:app-emission-prior-marginal}
\end{figure*}

\begin{figure*}[!tbp]
    \centering
    \includegraphics[width=0.92\textwidth]{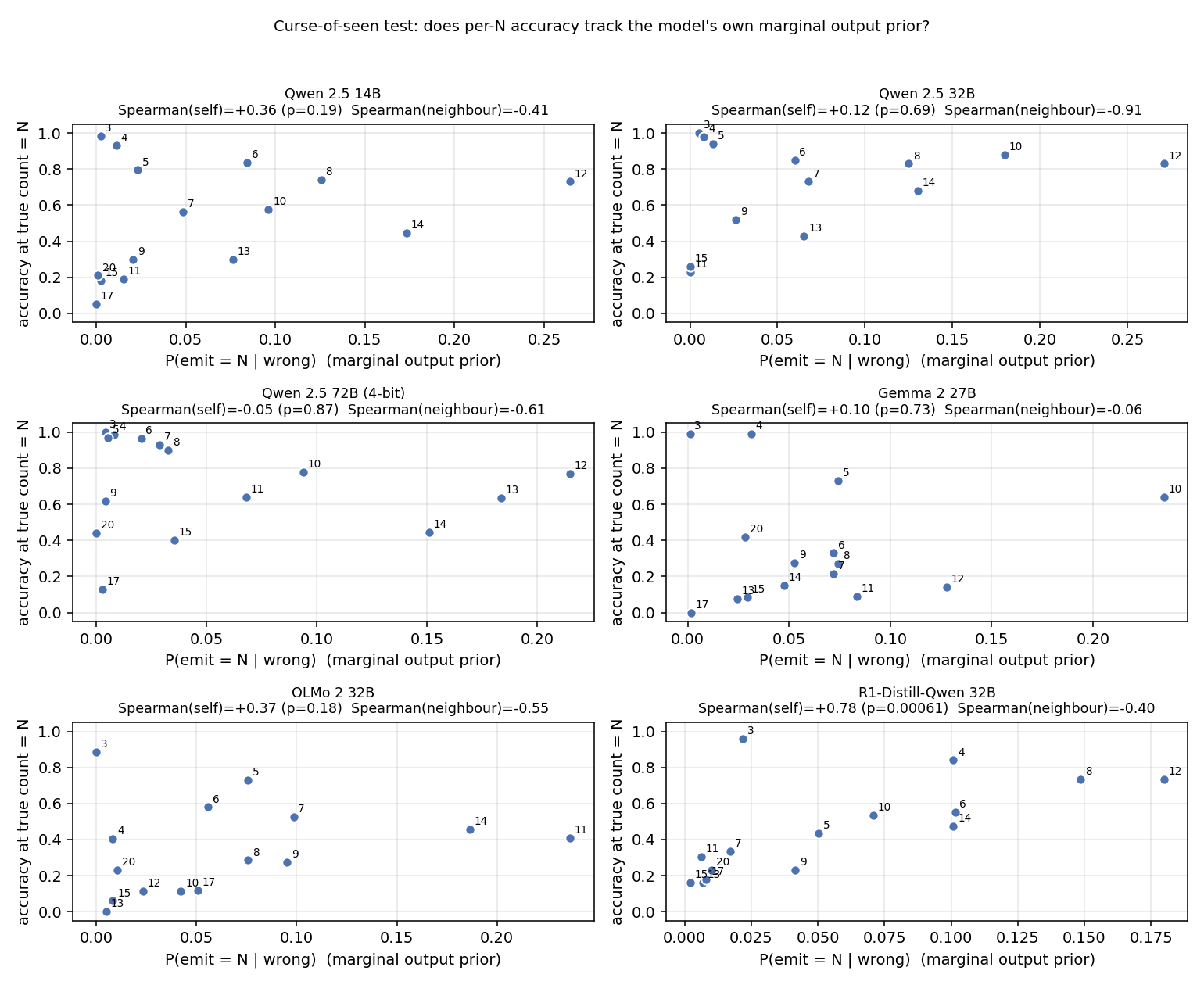}
    \caption{Accuracy versus model-output-prior proxy. For Qwen 32B and OLMo, per-count accuracy is lower when neighboring integers have high wrong-emission mass. This is a proxy for learned output prior, not a direct pretraining-frequency measurement.}
    \label{fig:app-emission-prior-correlation}
\end{figure*}

\section{Reproducibility}
\label{app:reproducibility}

This section records the seeds, hardware, and release plan needed to regenerate the main numbers.
Benchmark construction details are in Appendix~\ref{app:benchmark}; intervention protocols are in Appendix~\ref{app:interventions}.

\paragraph{Per-condition seeds.}
All sampling in the main and fine-grained evals uses \texttt{numpy.random.default\_rng(seed)} with the seeds in Table~\ref{tab:repro-seeds}.
The same seeds are used across all seven models, so the per-model accuracy grids cover identical $1{,}300$ + $200$ sequences.
The random-string pool of $200$ pseudowords is built once under seed $303$ and shared across both evals.

\begin{table}[h]
\centering\small
\begin{tabular}{lcc}
\toprule
condition & main eval seed & fine-grained seed \\
\midrule
\textsc{common\_words}        & $2$ & $12$ \\
\textsc{rare\_words}          & $3$ & $13$ \\
\textsc{multilingual\_single} & $4$ & $14$ \\
\textsc{multilingual\_mixed}  & $5$ & $15$ \\
\textsc{random\_strings}      & $6$ & $16$ \\
\midrule
random-string pool            & \multicolumn{2}{c}{$303$} \\
\bottomrule
\end{tabular}
\caption{Per-condition data-generation seeds. The same seeds are used across all seven models so each per-model accuracy grid is computed on the same $1{,}300$ + $200$ sequences. The random-string pool of $200$ pseudowords is built once under seed $303$ and shared across both evals.}
\label{tab:repro-seeds}
\end{table}

\paragraph{Compute.} All bf16 evaluations were run on a single NVIDIA H100 80GB. 4-bit evaluations (Qwen-72B and Llama-70B) used the same hardware. Approximate wall-clock per model: $\sim$$15$ minutes for the $1{,}300$-row main eval (non-reasoning), $\sim$$2$ hours for R1-Distill with thinking enabled (long generations), $\sim$$5$ minutes for the residual-capture pass that feeds the probe, $\sim$$2$ minutes for the per-layer probe fit on CPU. The full panel (seven models, main + fine eval + residuals) fits comfortably in a single working day on one H100. Mechanistic interventions on Qwen-32B (block decomposition, scale sweep over six $s$ values, feature-level patches) add $\sim$$2$ hours total.

\paragraph{Code and data release.}
Evaluation scripts, model stubs, probe and intervention drivers, raw result CSVs, and figure-generation code will be released upon acceptance.
Reproducing model evaluations requires accepting the original HuggingFace licenses for each checkpoint.

\end{document}